%% file: acl_latex.tex
\pdfoutput=1

\documentclass[11pt]{article}

\usepackage[preprint]{acl}

\usepackage{times}
\usepackage{latexsym}

\usepackage[T1]{fontenc}

\usepackage[utf8x,utf8]{inputenc}
\usepackage[russian,english]{babel} 
\newcommand\rus[1]{\begin{otherlanguage}{russian}{#1}\end{otherlanguage}}

\usepackage{microtype}

\usepackage{inconsolata}

\usepackage{graphicx}
\usepackage{dsfont}

\usepackage{enumitem}
\usepackage{rotating}

\usepackage{booktabs}
\usepackage{multirow}
\usepackage{subcaption}

\usepackage{arydshln}

\usepackage{CJKutf8}

\usepackage{tcolorbox}
\usepackage{amsthm}
\usepackage{amsmath}
\usepackage{amssymb}
\usepackage{xspace}
\newcommand{\narrowtextsc}[1]{\textls[-50]{\textsc{#1}}}
\newcommand{\lm}[1]{\texttt{#1}}
\newcommand{\sys}[1]{\narrowtextsc{#1}}
\newcommand{\data}[1]{\textsf{#1}}

\usepackage{tcolorbox}
\usepackage{xcolor,colortbl}

\definecolor{Gray}{gray}{0.94}
\definecolor{LightCyan}{rgb}{0.88,1,1}
\newcolumntype{a}{>{\columncolor{Gray}}c}
\newcolumntype{o}{>{\columncolor{white}}c}
\usepackage{soul}
\definecolor{celeste}{cmyk}{0.3922, 0.0353, 0, 0.1}
\definecolor{purple}{cmyk}{0.16, 0.28, 0, 0}
\definecolor{brilliantlavender}{cmyk}{0, 0.2235, 0, 0.1}
\definecolor{LightRed}{RGB}{232, 56, 107} 
\definecolor{LightBlue}{RGB}{116, 232, 226}
\definecolor{Tan}{rgb}{0.8203,0.7031,0.5469}
\definecolor{gblue}{RGB}{81,231,195}
\definecolor{orange}{RGB}{237, 183, 111}

\definecolor{greenblue}{RGB}{142, 207,201}

\definecolor{orange}{RGB}{255, 190, 122}

\definecolor{red}{RGB}{250, 127,111}

\definecolor{blue}{RGB}{130, 176, 210}

\usepackage{tikz}
\usepackage{array}
\usetikzlibrary{decorations.pathreplacing}
\newcommand{\roundedtable}[2]{
    \begin{tikzpicture}
        \node (table) [
            rectangle, 
            rounded corners, 
            draw=black,
            inner sep=0pt
        ] {
            \begin{tabular}{#1}
                #2
            \end{tabular}
        };
    \end{tikzpicture}
}

\usepackage{verbatim}

\usepackage{fontawesome5}

\title{Multilingual Datasets for Custom Input Extraction and Explanation Requests Parsing in Conversational XAI Systems}

\newcommand{\affilsup}[1]{\rlap{\textsuperscript{\normalfont#1}}}

\author{
    Qianli Wang\affilsup{1,2}
    \qquad
    Tatiana Anikina\affilsup{2,3}
    \qquad
    \textbf{Nils Feldhus\affilsup{1,2,5}}
    \qquad
    \textbf{Simon Ostermann\affilsup{2,3,4}}
    \\
    \textbf{Fedor Splitt\affilsup{1}}
    \quad
    \textbf{Jiaao Li\affilsup{1}}
    \quad
    \textbf{Yoana Tsoneva\affilsup{1}}
    \quad
    \textbf{Sebastian M\"oller\affilsup{1,2}}
    \quad
    \textbf{Vera Schmitt\affilsup{1,2}}
    \\
        $^1$Technische Universit\"at Berlin
    \quad
        $^2$German Research Center for Artificial Intelligence (DFKI) 
    \\
    $^3$Saarland Informatics Campus
    \quad
        $^4$Centre for European Research in Trusted AI (CERTAIN) 
    \\
        $^5$BIFOLD – Berlin Institute for the Foundations of Learning and Data
    \\
    \small{\textbf{Correspondence:} \texttt{\href{mailto:qianli.wang@tu-berlin.de}{qianli.wang@tu-berlin.de}}}
}

\begin{document}
\maketitle
\begin{abstract}
Conversational explainable artificial intelligence (ConvXAI) systems based on large language models (LLMs) have garnered considerable attention for their ability to enhance user comprehension through dialogue-based explanations. Current ConvXAI systems are often based on intent recognition to accurately identify the user's desired intention and map it to an explainability method. 
While such methods offer great precision and reliability in discerning users' underlying intentions for English, a significant challenge in the scarcity of training data persists, which impedes multilingual generalization. Besides, the support for free-form custom inputs, which are user-defined data distinct from pre-configured dataset instances, remains largely limited. To bridge these gaps, we first introduce \data{MultiCoXQL}, a multilingual extension of the \data{CoXQL} dataset spanning five typologically diverse languages, including one low-resource language. Subsequently, we propose a new parsing approach aimed at enhancing multilingual parsing performance, and evaluate three LLMs on \data{MultiCoXQL} using various parsing strategies. Furthermore, we present \data{Compass}, a new multilingual dataset designed for custom input extraction in ConvXAI systems, encompassing 11 intents across the same five languages as \data{MultiCoXQL}\footnote{Dataset and code are available at: \url{https://github.com/qiaw99/compass}}. We conduct monolingual, cross-lingual, and multilingual evaluations on \data{Compass}, employing three LLMs of varying sizes alongside \lm{BERT}-type models.\looseness=-1
\end{abstract}

\section{Introduction}
To improve the transparency of LLMs while ensuring efficiency and user comprehension, conversational XAI systems have recently emerged \cite{chromik-butz-2021-human-xai-interaction, lakkaraju-2022-rethinking, shen-2023-convxai, bertrand-2023-selective, mindlin2024measuring, feustel-etal-2024-enhancing, he2025conversational}. Several systems have since been developed, e.g., \sys{TalkToModel} \cite{slack-2023-talktomodel}, \sys{InterroLang} \cite{feldhus-etal-2023-interrolang} and \sys{LLMCheckup} \cite{wang-etal-2024-llmcheckup}. These systems include user interfaces that facilitate users to interact in natural language with a system and rely on intent recognition\footnote{Intent recognition and parsing are used interchangeably in the scope of this work.}. Intent recognition is a key upstream component in ConvXAI systems, focusing on accurately interpreting user inputs from multiple perspectives \cite{chen-etal-2022-unidu} and mapping user intents to the corresponding explainability methods, which enables explanations that are as faithful as the underlying method allows \cite{wang-etal-2024-coxql}. 
Nevertheless, intent recognition remains challenging for ConvXAI, due to the scarcity of training data, particularly multilingual data, and the specialized nature of the XAI domain, which involves mapping requests across a diverse range of XAI methods. \citet{wang-etal-2024-coxql} presents the first and, to-date, largest dataset, \data{CoXQL}, for intent recognition in ConvXAI, but it is limited to English. Entering queries in other languages may either result in undesired explanations or prevent users from fully accessing the ConvXAI capabilities, highlighting the need to extend intent recognition to multiple languages for effective use in multilingual scenarios \cite{gerz-etal-2021-multilingual,shi-etal-2022-xricl}. \looseness=-1

\begin{figure*}[t!]
\centering
\resizebox{\textwidth}{!}{
\begin{minipage}{\columnwidth}
\includegraphics[width=\columnwidth]{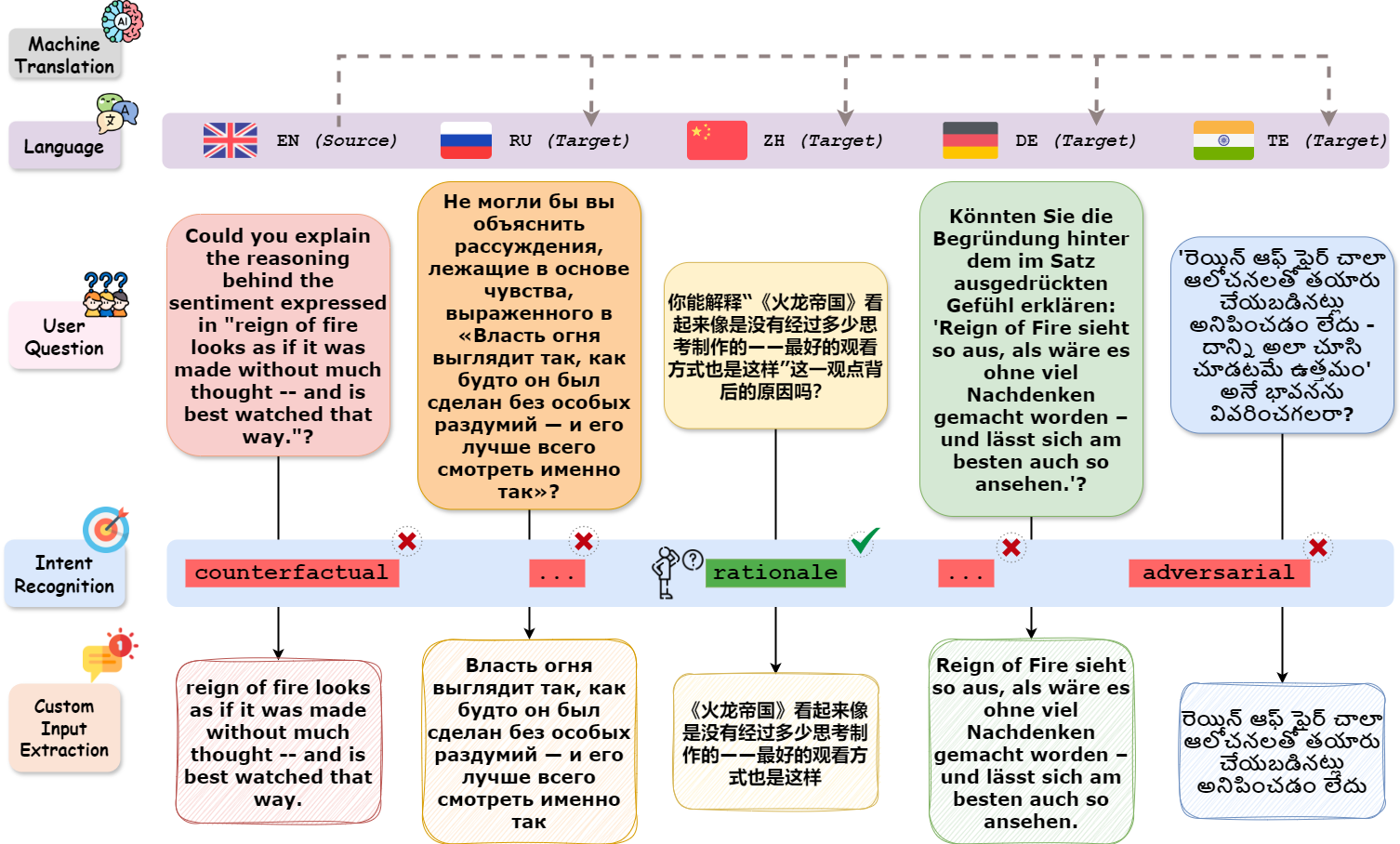}
\end{minipage}
}
\caption{Example parallel utterances from \data{Compass}, including the user question, corresponding intent, and extracted custom input, are provided in \textit{English} (EN), \textit{Russian} (RU), \textit{Chinese} (ZH), \textit{German} (DE), and \textit{Telugu} (TE). The example question is requesting reasoning (parsed as ``\textit{rationale}'', which should provide a natural language explanation) and the corresponding extracted custom input comes from the \data{SST2} dataset.}
\label{fig:compass_example}
\end{figure*}


Furthermore, \data{CoXQL} is restricted to an existing dataset with fixed data points (\S\ref{sec:compass}) that can be queried by the user and explained, as in \sys{TalkToModel}, \sys{InterroLang}, and \sys{LLMCheckup}. The deficiency is that users are unable to freely explore custom input based on their preferences with the explained LLMs, which hinders the generalizability of ConvXAI systems.  One goal of ConvXAI systems is to support more adaptive usage (\S\ref{sec:compass}), allowing personalization \cite{orji-2017-personalization}, user engagement \cite{irfan-2019-engagement}, and efficient use of systems \cite{BURKOLTER2014346}. This can be realized by allowing custom input (Figure~\ref{fig:compass_example}). Nonetheless, the current custom input extraction in the context of ConvXAI has been largely constrained by the lack of suitable datasets. \looseness=-1

\begin{figure*}[t!]
\centering
\resizebox{\textwidth}{!}{
\begin{minipage}{\columnwidth}
\includegraphics[width=\columnwidth]{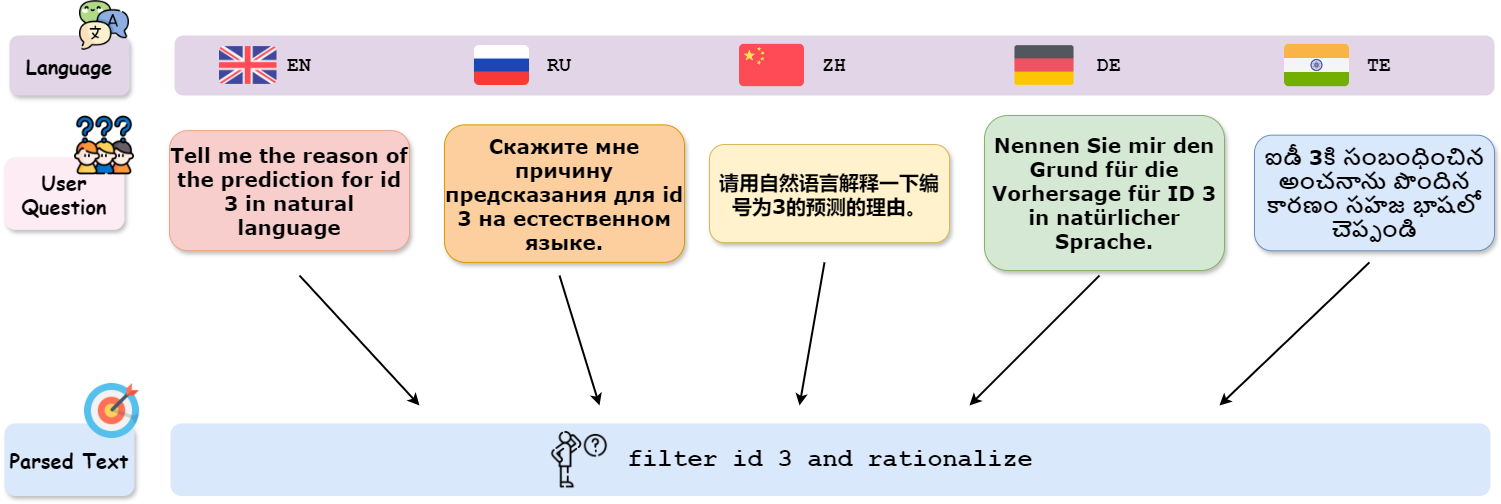}
\end{minipage}
}
\caption{Example parallel utterances from \data{MultiCoXQL}, including the user question, and corresponding parsed texts, are provided in \textit{English} (EN), \textit{Russian} (RU), \textit{Chinese} (ZH), \textit{German} (DE), and \textit{Telugu} (TE). The example question seeks reasoning (parsed as ``\textit{rationale}'') for a specific instance with the id 3, which should return a natural language explanation.}
\label{fig:multicoxql_example}
\end{figure*}

To address these gaps, we \textbf{first} extend \data{CoXQL} \cite{wang-etal-2024-coxql} to support multiple languages, called \data{MultiCoXQL} (Figure~\ref{fig:multicoxql_example}). \data{MultiCoXQL} is created by machine translating \data{CoXQL} instances, while preserving their intent annotations, covering 5 languages: \textit{German}, \textit{Chinese}, \textit{Russian}, \textit{Telugu} and \textit{English}. We assess the quality of machine translation through human annotators who assess meaningfulness and correctness. The Chinese and German translations are of high quality (Figure~\ref{fig:translation_quality}), and the similarity between the original English text and the translated text improves after human correction. \textbf{Secondly}, we evaluate the effectiveness of one baseline and three state-of-the-art parsing approaches in ConvXAI on \data{MultiCoXQL}. To improve upon the limited cross-lingual generalization of existing approaches, we propose \underline{G}uided \underline{M}ulti-prompt \underline{P}arsing (GMP), which combines existing methods and noticeably enhances multilingual parsing accuracy. \looseness=-1
\textbf{Thirdly}, we present the \data{Compass} dataset\footnote{Abbreviation of ``\underline{C}ust\underline{om} Input Extraction and Explanation Requests \underline{Pa}rsing in ConvXAI \underline{S}ystem\underline{s}'' (\data{Compass}).} (Figure~\ref{fig:compass_example}, Figure~\ref{fig:ops}) for enabling custom input in ConvXAI. It includes user questions, extracted custom inputs, and corresponding intents across the five aforementioned languages. We conduct monolingual, cross-lingual, and multilingual evaluations on \data{Compass} using \lm{(m)BERT} and three decoder-only LLMs. For intent recognition, fine-tuned \lm{BERT} performs comparably to that of the LLMs and outperforms them in Chinese, German, and Telugu. For custom input extraction, out of four approaches, GOLLIE \cite{sainz2024gollie} performs best with smaller LLMs, while na\"ive few-shot prompting presents the best results with larger LLMs. 


\section{Related Work}
\paragraph{Parsing in ConvXAI Systems} In most prior ConvXAI systems \cite{werner-2020-eric, nguyen-2023-xagent, shen-2023-convxai}, parsing is achieved by comparing the semantic similarity between user queries and a predefined set of example utterances, often resulting in relatively low parsing accuracy. In contrast, \sys{TalkToModel} \cite{slack-2023-talktomodel} converts user questions into SQL-like queries for parsing and employs guided decoding (GD) by defining a grammar to constrain the output vocabulary. Similarly, \sys{InterroLang} \cite{feldhus-etal-2023-interrolang} uses fine-tuned models for slot tagging to perform parsing, while \citet{wang-etal-2024-llmcheckup} introduces multi-prompt parsing (MP), which hierarchically parses user intent from coarse-grained to fine-grained slots. To ensure compatibility with various XAI operations, MP with template checking (MP+) \cite{wang-etal-2024-coxql} is proposed to validate the output and fulfill operational requirements. Our work further investigates the effectiveness of GD, MP, and MP+ in multilingual settings and proposes a new approach, Guided Multi-prompt Parsing, which substantially enhances multilingual parsing performance.

\paragraph{Multilingual Dataset Annotation} \data{XNLI} \cite{conneau-etal-2018-xnli} extend \data{MultiNLI} \cite{williams-etal-2018-broad}  to 15 languages, including low-resource languages, to facilitate cross-lingual natural language inference. \citet{min-etal-2019-pilot}, \citet{tuan-nguyen-etal-2020-pilot}, and \citet{bakshandaeva-etal-2022-pauq} broaden \data{Spider} dataset \cite{yu-etal-2018-spider}, a widely recognized text-to-SQL dataset in English that encompasses queries of varying complexity, by translating it into target languages. \data{MultiSpider} is subsequently created as a multilingual Text-to-SQL dataset, covering seven popularly used languages \cite{Dou-2022-MultiSpiderTB}. \citet{hennig-etal-2023-multitacred} introduce the \data{MultiTACRED} dataset, created by translating \data{TACRED} \cite{zhang-etal-2017-position}, a dataset for information extraction, into 12 typologically diverse languages from nine language families. Our work is closely aligned with prior research on multilingual dataset creation and annotation, and adheres to established best practices in the field.



\paragraph{Information Extraction}
Information extraction can be tackled using in-context learning, which leverages the emergent capability of LLMs \cite{han2024empiricalstudyinformationextraction}. 
We rely on information extraction approaches to identify custom user inputs in ConvXAI systems. \sys{TANL} translates between input and output text using an augmented natural language format, with the output later decoded into structured objects \cite{paolini2021structured}. \sys{GPT-NER} reformulate information extraction as a sequence-to-sequence task and special tokens are used to demarcate the boundaries of extracted entities \cite{wang2023gptnernamedentityrecognition}. \sys{GOLLIE} employs annotation guidelines represented in code snippet for both input and output \cite{sainz2024gollie}. This approach is effective when the extracted entities can be represented in a structured or code-like format. In our paper, we employ these approaches to evaluate their efficacy in capturing custom input within ConvXAI systems and across multilingual settings.\looseness=-1


\section{The \data{MultiCoXQL} Dataset}
\label{subsec:automatic_translation}

\data{CoXQL} \cite{wang-etal-2024-coxql} is a text-to-SQL dataset for intent recognition in ConvXAI systems and comprises \textit{user questions} and \textit{gold labels} (SQL-like queries) in English. \data{CoXQL} covers 31 operations, including explainability and supplementary operations\footnote{Appendix~\ref{app:coxql} includes details on operations and examples.} (Table~\ref{tab:coxql_ops}). Some operations involve additional fine-grained slots\footnote{E.g., the feature importance method can support various approaches, including \textit{LIME}, \textit{Input x Gradient}, \textit{Integrated Gradients}, and \textit{attention} in \data{CoXQL}. The number of data points for each operation depends on whether it involves additional slots.} and multiple interpretations of the same request, rendering intent recognition in ConvXAI particularly challenging. \looseness=-1


The \data{MultiCoXQL} dataset introduced in this work encompasses five languages: \textit{English}, \textit{German}, \textit{Russian}, \textit{Chinese}, and \textit{Telugu} (Figure~\ref{fig:multicoxql_example}). These languages are selected for their typological diversity, representing a spectrum from widely spoken to low-resource languages that use different scripts. 
Following \citet{hennig-etal-2023-multitacred} and \citet{popov2024donortransfernlpdatasets}, we translate the entire train and test splits of \data{CoXQL} into the target languages using \lm{Gemini-1.5-pro}\footnote{The prompt instruction to translate user texts into target languages is provided in Appendix~\ref{app:prompt_instruction}. \lm{Gemini-1.5-pro} is selected, as it supports all target languages that we determine: \url{https://ai.google.dev/gemini-api/docs/models/gemini\#available-languages}.} \cite{geminiteam2024gemini15unlockingmultimodal} and name it \data{MultiCoXQL} (Figure~\ref{fig:multicoxql_example}), whose translation quality is evaluated in \S\ref{subsec:translation_evaluation}. Note we only translate the user questions, not the gold labels, in order to maintain consistency in the label space. Finally, we store the translated instances in the same JSON format as the original \data{CoXQL} English dataset.

\section{The \data{Compass} Dataset}
\label{sec:compass}
Custom input (Figure~\ref{fig:compass_example}) refers to user-defined or task-specific data provided to ConvXAI, distinct from instances found in pre-configured datasets (Figure~\ref{fig:multicoxql_example}). In previous ConvXAI systems, users can only query instances from pre-configured datasets using their dataset ID, while the main challenge in custom input is for LLMs to explicitly interpret and extract relevant information from the user's question. Custom input allows users to explore ConvXAI systems according to their individual preferences, thereby enhancing system \textit{flexibility}, \textit{generalizability}, and \textit{extensibility} \cite{BURKOLTER2014346, orji-2017-personalization}. However, no publicly available dataset currently addresses this type of input for information extraction tasks within ConvXAI systems. \data{Compass} is therefore constructed to address this gap and serves as a \textbf{synergistic complement} to \data{(Multi)CoXQL}, offering users diverse input text formats. To maintain consistent language selection, \data{Compass} adopts the same five languages as \data{MultiCoXQL} (\S\ref{subsec:automatic_translation}). \looseness=-1

\begin{figure}[t!]
\centering
\resizebox{\columnwidth}{!}{
\begin{minipage}{\columnwidth}
\includegraphics[width=\columnwidth]{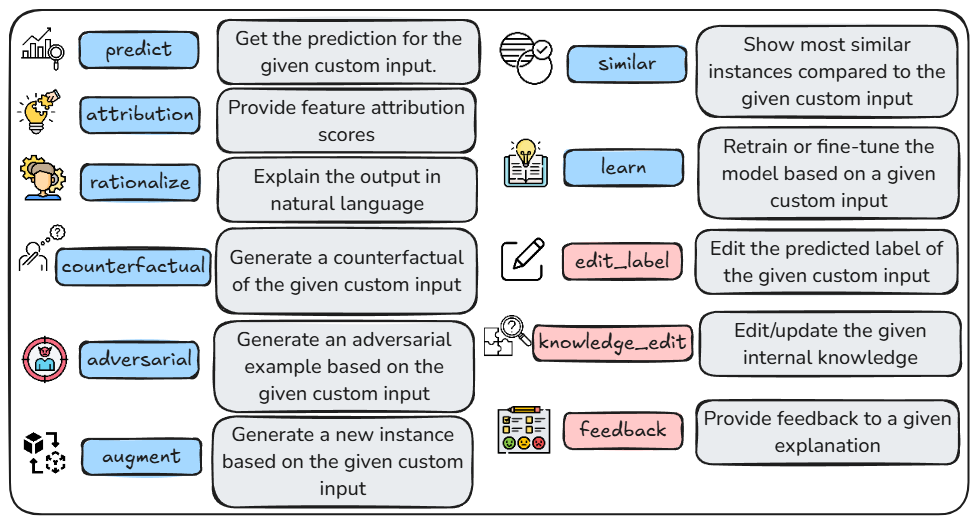}
\end{minipage}
}
\caption{Main operations in the \data{Compass} dataset capable of receiving custom input. Operations highlighted in $\color{blue}{blue}$ are collected from the \data{CoXQL} dataset. Operations highlighted in $\color{red}{red}$ are not yet implemented in any ConvXAI system and have been identified from the literature.}
\label{fig:ops}
\end{figure}

\subsection{Operations}
\label{subsec:compass_ops}
As shown in Figure~\ref{fig:ops}, we first identify eight operations within \data{CoXQL} that should accommodate custom user input to enhance user experience and these operations are highlighted in blue. Additionally, we include three new operations (\texttt{edit\_label}, \texttt{knowledge\_edit}, \texttt{feedback}), highlighted in red, drawn from the literature \cite{li-etal-2022-using, zhang-etal-2024-knowledge-editing,wang-etal-2025-cross}. These operations provide users with deeper insights into model behavior by enabling more interactive engagement with the underlying explained model.

\subsection{Dataset Construction}
\label{subsec:dataset_construction}
\paragraph{Source of Custom Input} To preserve the naturalness and usability, we curate custom input from three core NLP tasks - fact-checking, commonsense question answering, sentiment analysis - ensuring all examples remain self-contained within the NLP domain. This approach endows \data{Compass} with diverse topics and texts of different complexity and length, offering varying levels of difficulty for LLMs in custom input extraction. The following datasets are selected for each use case: \data{COVID-Fact} \cite{saakyan-etal-2021-covid},  \data{ECQA} \cite{aggarwal-etal-2021-explanations}, and \data{SST2} \cite{socher-etal-2013-recursive}\footnote{\data{COVID-Fact} is a fact-checking dataset consisting of claims and evidence, with labels indicating whether a claim is \textit{supported} or \textit{refuted}. \data{ECQA} is commonsense question answering dataset encompassing commonsense questions with multiple-choice answers. \data{SST2} is a sentiment analysis dataset which provides movie reviews and corresponding sentiment labels. Examples from each dataset are shown in Figure~\ref{fig:samples}.}.

\paragraph{Dataset Creation} Data points in \data{Compass} consist of a  \textit{user question}, \textit{custom input} and the \textit{corresponding intent}. As a first step, we manually create $10$ examples per operation in English, similar to the ones in Figure~\ref{fig:compass_example}, following the approach used in \citeposs{feldhus-etal-2023-interrolang} study to simulate how questions may naturally arise. This ensures alignment with realistic scenarios while maintaining structural and stylistic diversity across examples. These manually created examples are then used as demonstrations to prompt \lm{Gemini}, guiding it to generate new data points that conform to the desired question formulations and linguistic patterns. The resulting user questions with the custom input are evaluated by checking if they directly pertain to the specified operation and discarded if not applicable. Furthermore, we manually develop a test set adhering to the guidelines (Figure~\ref{fig:testset}), incorporating questions for each included operation (Figure~\ref{fig:ops}). Ultimately, we acquire a \textbf{training set} comprising 1089 instances and a \textbf{test set} consisting of 109 instances. \looseness=-1

\begin{figure}[t!]
\centering
\resizebox{\columnwidth}{!}{
\begin{minipage}{\columnwidth}
\includegraphics[width=\columnwidth]{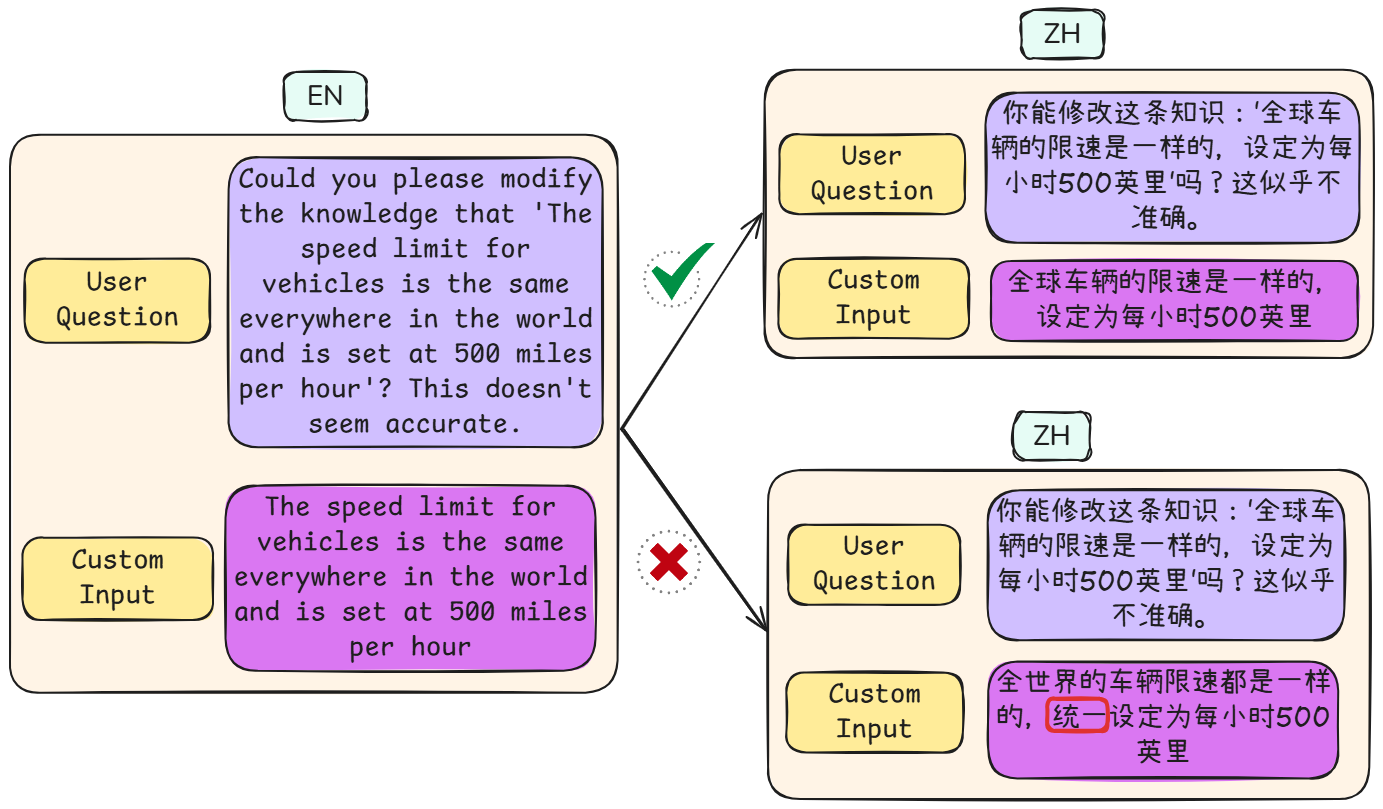}
\end{minipage}
}
\caption{Validation of whether the translated custom input (e.g., in \textit{Chinese}) is fully contained within the translated user question. The words marked in the red box are not included in the translated user question, and thus the translation in the bottom right is invalid.}
\label{fig:validation}
\end{figure}


\subsection{Automatic Translation}
Consistent with the translation process for \data{MultiCoXQL} (\S\ref{subsec:automatic_translation}), we prompt \lm{Gemini} (Figure~\ref{fig:translation_prompt}) to translate both the \textit{user question} and the \textit{custom input} into the target language, with the translation quality subsequently evaluated in \S\ref{subsec:translation_evaluation}. We then verify that the translated custom input remains fully embedded within the translated user question (Figure~\ref{fig:validation}); if not, the translation process is repeated until a valid result is obtained.

\section{Methodology}


\subsection{\data{MultiCoXQL}}
\label{subsec:methodology_multicoxql}
In \data{CoXQL}, the recognition of XAI intents is treated as a task similar to text-to-SQL (Figure~\ref{fig:coxql_example}), which can be represented and processed as a sequence-to-sequence task \cite{sutskever-2014-seq2seq}. In this work, we benchmark one baseline and three state-of-the-art parsing approaches (\S\ref{subsubsec:prior_work}) on \data{MultiCoXQL}, and propose a new method (\S\ref{subsubsec:gdp}) for explanation request parsing in multilingual settings.

\subsubsection{Parsing Approaches Selection}
\label{subsubsec:prior_work}




\textbf{Nearest neighbor} (NN) determines intents based on the semantic similarity between the user query and existing training samples, using a multilingual \lm{SBERT} model\footnote{\url{https://huggingface.co/sentence-transformers/paraphrase-multilingual-MiniLM-L12-v2}}, which is trained on all target languages (\S\ref{subsec:automatic_translation}). \textbf{Guided decoding} (GD) ensures that the output conforms to predefined grammatical rules and constraints \cite{shin-etal-2021-constrained}. \textbf{Multi-prompt parsing} (MP) \cite{wang-etal-2024-llmcheckup} comprises two stages: first, the model is presented with all possible operations in a simplified format to identify the main operation; subsequently, the model is further prompted to populate fine-grained attributes (\S\ref{subsec:automatic_translation}). Unlike GD, MP is not constrained by grammatical rules and thus tends to deviate from the predefined templates for each operation. \textbf{Multi-prompt parsing with template checking} (MP+) addresses this issue to some extent by incorporating template validation \cite{wang-etal-2024-coxql}. 

\subsubsection{Guided Multi-prompt Parsing} 
\label{subsubsec:gdp}
To leverage and integrate the strengths of GD and MP, particularly in multilingual settings where existing methods often yield suboptimal performance (Table~\ref{tab:multi_coxql_parsing}), we propose a simple yet effective approach: Guided Multi-prompt Parsing (GMP) (Figure~\ref{fig:guided_mp_workflow}). First, we employ \lm{SBERT} to compute intent centroid embeddings by averaging the embeddings of training examples that share the same intent. We then find which intents are most similar to the user's query using cosine similarity between the intent centroid embeddings and the user query embedding, and retrieve the top-$k$ most similar training examples for each candidate intent based on their similarity to the user input. These retrieved examples are then used to dynamically construct a prompt for generating a coarse-grained intent (e.g., \textit{learn} or \textit{augment}) that corresponds to the supported XAI operations. Next, GMP uses prompting with an intent-specific grammar, guided decoding and additional demonstrations to generate a fine-grained intent with relevant attributes (Table~\ref{tab:coxql_ops}). The multi-stage prompting with multiple intent options provides greater flexibility, while guided decoding ensures that the final structured parse includes the only correct associated attributes\footnote{Further details on GMP can be found in Appendix \ref{app:guided_mp}.}.

\subsection{\data{Compass}}
\subsubsection{Intent Recognition}
\label{subsubsec:intent_recognition}
\data{Compass} embodies coarser-grained intents compared to \data{(Multi)CoXQL}, with a pronounced focus on custom input extraction (Figure~\ref{fig:compass_example}). For \lm{(m)BERT}, we frame intent recognition as a multi-class classification task and fine-tune \lm{(m)BERT} on the training dataset for a given language. In addition, in-context learning is employed for decoder-only LLMs (\S\ref{sec:model}) to perform intent recognition. Suitable demonstrations are selected based on semantic similarity, measured using a multilingual \lm{SBERT}, followed by the application of few-shot prompting.


\subsubsection{Information Extraction}
\label{subsubsec:custom_input_extraction}
To facilitate custom input extraction from user requests, we formulate the task as a \textit{sequence labeling} problem, where the custom input embedded within the user request is treated as the target output. We consider four distinct information extraction approaches for identifying custom inputs\footnote{Prompts for each of the following four approaches are provided in Appendix~\ref{app:custom_input_prompt}.}. 

\paragraph{Na\"ive} For \lm{(m)BERT}, custom input extraction is framed as a token-level classification task, whereas for decoder-only LLMs (\S\ref{sec:model}), few-shot prompting is performed using $n=10$ demonstrations (\S\ref{subsubsec:intent_recognition}).

\paragraph{TANL} \texttt{TANL} \cite{paolini2021structured} employs predefined inline tagging to annotate entities, thereby capturing structural information within the text.

\paragraph{GPT-NER} \texttt{GPT-NER} \cite{wang2023gptnernamedentityrecognition} reformulates sequence labeling as a text generation problem, where the model generates augmented text with information marked with special tokens. 

\paragraph{GOLLIE} \texttt{GOLLIE} \cite{sainz2024gollie} leverages annotation guidelines to guide the model, providing detailed instructions on how to annotate specific types of information.





\section{Models}
\label{sec:model}
We select three open-source, state-of-the-art decoder-only LLMs with increasing parameter sizes from distinct model families: \lm{Llama3-8B} \cite{llama3modelcard}, \lm{Phi4-14B} \cite{abdin2024phi4technicalreport}, and \lm{Qwen2.5-72B} \cite{qwen2024qwen25technicalreport} to evaluate the \data{MultiCoXQL} and \data{Compass} datasets. These models are selected because they have been trained on at least one of the non-English target languages we chose (\S\ref{subsec:automatic_translation}). In contrast to few-shot prompting, for fine-tuning, we employ \lm{BERT} and \lm{mBERT} models \cite{devlin-etal-2019-bert} to conduct monolingual, cross-lingual, and multilingual evaluations (\S\ref{subsec:automatic_evaluation})\footnote{The details regarding LLMs and language-specific pre-trained \lm{(m)BERT} models are listed in Appendix~\ref{app:model}.}.\looseness=-1

\section{Evaluation}

\subsection{Machine Translation Human Evaluation}
\label{subsec:translation_evaluation}
To evaluate the quality of machine translation, we engage 8 in-house native speakers of each target language to meticulously review the translations\footnote{Instructions for human evaluation and details about the annotators' background are provided in Appendix~\ref{app:human_evaluation}.}, rectify the translations if necessary, and assess their quality by answering two questions: 

\begin{itemize}[noitemsep,topsep=0pt,leftmargin=*]
\item (Q1) \textit{``Does the translated text effectively convey the semantic meaning of the English original, despite minor translation errors?''} 
\item (Q2) \textit{``Is the overall translation grammatically correct?''}
\end{itemize}

In addition, we leverage the multilingual \lm{SBERT} (\S\ref{subsec:methodology_multicoxql}), to measure the semantic similarity between the original input and its translations before and after human evaluation. 


\begin{figure}[t!]
\centering
\resizebox{\columnwidth}{!}{
\begin{minipage}{\columnwidth}
\includegraphics[width=\columnwidth]{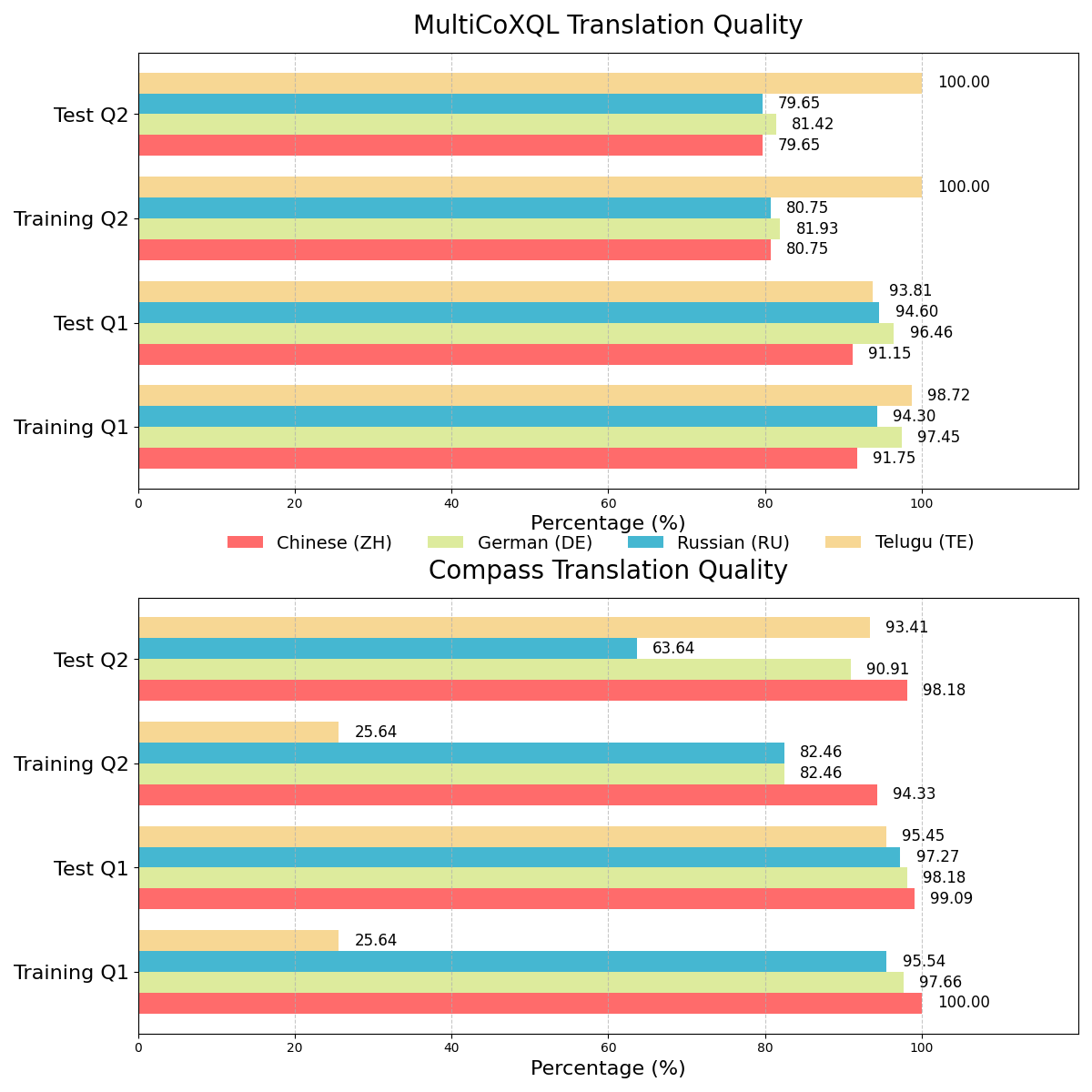}
\end{minipage}
}
\caption{Translation quality of texts in Chinese (ZH), German (DE), Telugu (TE) and Russian (RU) from the training and test sets of \data{MultiCoXQL} and \data{Compass}, as judged by native speakers.}
\label{fig:translation_quality}
\end{figure}

\subsection{Automatic Evaluation}
\label{subsec:automatic_evaluation}
To evaluate the capability of models in interpreting user intents, we measure the performance of three LLMs (\S\ref{sec:model}), ranging in size from 8B to 72B, using five approaches: NN, GD, MP, MP+, and GMP (\S\ref{subsec:methodology_multicoxql}). Intent recognition performance is evaluated by $F_1$ score on \data{MultiCoXQL} and on \data{Compass}. In parallel, we evaluate the same LLMs for custom input extraction on \data{Compass} using four approaches: Na\"ive, TANL, GPT-NER, and GOLLIE (\S\ref{subsubsec:custom_input_extraction}). Information extraction performance is likewise reported using $F_1$ scores. Additionally, we employ \lm{BERT} and \lm{mBERT} (Table~\ref{tab:bert_models}) for monolingual, cross-lingual, and multilingual evaluation of both intent recognition and information extraction on the \data{Compass} dataset.

\paragraph{Monolingual} We utilize a \lm{BERT} model pre-trained on the target language and fine-tune it using the training set in the given language. 

\paragraph{Cross-lingual} We evaluate the performance of a multilingual \lm{mBERT} model on the test set of each of the four target languages (\S\ref{subsec:automatic_translation}), as well as English, after training it only on the English training set.  

\paragraph{Multilingual} Simultaneously, we train a multilingual \lm{mBERT} model on a mixed dataset comprising all languages. \lm{mBERT} is trained on the full English training split along with a variable proportion of the target language's training split, as proposed by \citet{nag-etal-2021-data}. We vary the amount of target language data used to \{10\%, 25\%, 50\%, 75\%, 100\%\} of the available training set.

\input{table/multicoxql_parsing.tex}

\section{Results and Analysis}
\subsection{Machine Translation Evaluation}

Figure~\ref{fig:translation_quality} illustrates the translation quality  for \data{MultiCoXQL} and \data{Compass} across all selected languages, highlighting that \lm{Gemini} performs well overall. Notably, translations into Chinese and German are of relatively high quality compared to Telugu, particularly on \data{Compass}. Telugu translations occasionally pose challenges for \lm{Gemini}, largely due to the semantic complexity of the custom input and language's low-resource status. In addition, as shown in Table~\ref{tab:similarity}, the translated texts generally exhibit a high degree of similarity to the original English input. Among the target languages, the German texts have the highest similarity scores, whereas the Telugu texts demonstrate the lowest. Moreover, after human annotators revise the translations, the similarity improves by up to $9\%$\footnote{Further quality analysis and common error patterns are detailed in  Appendix~\ref{app:quality_analysis}. Given the challenges associated with recruiting multiple annotators—particularly for low-resource languages—and the relatively straightforward nature of the task for human annotators, we report inter-annotator agreement (IAA) only for the German and Chinese test sets, achieving Krippendorff’s $\alpha$ scores of 0.89 and 0.94, respectively.}.\looseness=-1

\subsection{\data{MultiCoXQL}}
\input{table/monolingual_intent_recognition}

Table~\ref{tab:multi_coxql_parsing} reveals that MP and MP+ outperform GD on the English subset of \data{MultiCoXQL}, consistent with the findings of \citet{wang-etal-2024-llmcheckup, wang-etal-2024-coxql}. All three approaches significantly surpass the baseline. However, GD generally exhibits superior parsing performance in other languages, especially in Chinese and Telugu, compared to MP and MP+. This discrepancy can be attributed to the limited cross-lingual generalizability of current methods  (\S\ref{subsec:methodology_multicoxql}). For Telugu, GD achieves performance on par with other languages, whereas MP and MP+ exhibit a marked performance decline. Moreover, due to the hierarchical, two-stage parsing nature of MP and MP+, coupled with their lack of grammatical constraints compared to GD, they are more prone to misidentifying the main operation or generating output that fall outside the predefined operation set, thereby hampering further parsing. This issue is partially addressed by our proposed approach, GMP, which performs two-stage parsing similar to MP(+), while constraining the outputs using predefined grammars. As shown in Table~\ref{tab:multi_coxql_parsing}, GMP consistently outperforms existing methods by an average of 28.31\%, particularly demonstrating substantial performance gains across non-English languages and achieves comparable performance on \lm{Qwen2.5-72B} and \lm{Phi4-14B}, both of which significantly outperform \lm{Llama3-8B}. Meanwhile, in English, GMP occasionally underperforms MP(+). This can be attributed to the application of grammars, which limit the flexibility of generation, while ensuring the outputs conforms to predefined grammatical structures.

\input{table/multilingual_intent_recognition}

\subsection{\data{Compass}}
\subsubsection{Intent Recognition}

\paragraph{Monolingual Evaluation} Table~\ref{tab:monoligual_intent_recognition} illustrates that while LLMs achieve satisfactory accuracy on English data, they struggle to recognize user intents in the Chinese and Telugu subsets of \data{Compass}. Model performance generally improves with increasing model size. Furthermore, fine-tuned \lm{BERT} achieves performance comparable to \lm{Qwen2.5-72B}, and consistently outperforms \lm{Llama3-8B} and \lm{Phi4-14B} in Chinese, German and Telugu, offering an efficient solution for intent recognition. We observe that LLMs occasionally generate labels in the target language instead of English (Figure~\ref{fig:incorrect_example_intent}).
\looseness=-1

\paragraph{Cross-lingual \& Multilingual Evaluation} Table~\ref{tab:multiligual_intent_recognition} shows cross-lingual \lm{mBERT} yields lower performance compared to monolingual \lm{BERT}, whereas  multilingual \lm{mBERT} consistently outperforms both. Moreover, in the multilingual setting, performance improves as the proportion of non-English training data increases ($\Delta$), with especially compelling performance gains observed for Telugu. 

\input{table/custom_input_extraction}
\subsubsection{Custom Input Extraction}
\paragraph{Monolingual Evaluation} 
Table~\ref{tab:custom_input_extraction} unveils that extracting custom input in Telugu poses rigorous challenges, with none of the evaluated approaches or models achieving adequate results. For \lm{Llama3-8B} and \lm{Phi4-14B}, GOLLIE generally outperforms the na\"ive approach, GPT-NER, and TANL, most notably in German and Russian, where the performance margin is substantial (with \lm{Llama3-8B} on Chinese, performance improves by up to $200\%$ when comparing GOLLIE to na\"ive prompting). In contrast, for \lm{Qwen2.5-72B}, the na\"ive approach yields the best results among all considered methods. On \data{Compass}, smaller LLMs benefit from GOLLIE, which reformulates the task into structured code snippets, making it more interpretable for models \cite{sainz2024gollie}. Conversely, larger models appear more susceptible to distraction from newly introduced patterns (Figure~\ref{fig:first_prompts}, Figure~\ref{fig:second_prompts}). In addition, fine-tuned \lm{BERT} exhibits competitive performance across all target languages. For non-English subsets, \lm{BERT} generally outperforms LLMs across most approaches, particularly in Telugu, where it achieves more than double the performance of \lm{Qwen2.5-72B} using the na\"ive approach.

\paragraph{Cross-lingual \& Multilingual Evaluation} 

As shown in Table~\ref{tab:multiligual_custom_input_extraction}, cross-lingual \lm{mBERT} exhibits lower performance compared to monolingual \lm{BERT}, aligned with the results observed in the intent recognition task, in particular with a pronounced performance gap in Telugu. As the proportion of non-English data increases, the performance improvement trend is similar to that shown in Table~\ref{tab:multiligual_intent_recognition}, with Telugu benefiting the most. Besides, multilingual \lm{mBERT} consistently outperforms all LLMs across nearly all languages, with the exception of English.

\input{table/multilingual_custom_input_extraction}

\paragraph{Error Analysis} Figure~\ref{fig:incorrect_example_custom_input} illustrates common error patterns observed in the custom input extraction outputs from LLMs. In some cases, LLMs tend to generate or substitute words that do not appear in the original user question, extract only part of the intended custom input, or inadvertently include parts of artifacts from extraction methods in the final output. Additionally, there are instances where LLMs fail to solve the task altogether due to the task's inherent difficulties for LLMs.

\section{Conclusion}
In this work, we first extend the \data{CoXQL} dataset for intent recognition in ConvXAI to a multilingual version, \data{MultiCoXQL}, covering five languages, including one low-resource language, using machine translation followed by human evaluation and correction. We benchmark state-of-the-art explanation request parsing approaches on \data{MultiCoXQL} using three different LLMs. 
Second, we propose a new approach, Guided Multi-Prompt Parsing, which integrates the strengths of existing methods and substantially improves parsing accuracy in multilingual settings. Third, we introduce the \data{Compass} dataset for coarse-grained intent recognition and custom input extraction in ConvXAI, incorporating the same five languages as \data{MultiCoXQL}. We conduct comprehensive experiments using three LLMs, along with \lm{(m)BERT}, on \data{Compass}, to evaluate performance in monolingual, cross-lingual, and multilingual scenarios. We observe that cross-lingual \lm{mBERT} underperforms compared to monolingual \lm{mBERT}, whereas multilingual \lm{mBERT} outperforms both. For the task of custom input extraction, GOLLIE proves to be more effective for smaller LLMs, while na\"ive few-shot prompting yields better results with larger LLMs.


\section*{Limitations}
A key limitation of this work is its dependence on a machine translation (MT) system, i.e., \lm{Gemini-1.5-pro}, to obtain high-quality translations for the \data{MultiCoXQL} and \data{Compass} datasets. Depending on the availability of linguistic resources and the quality of the MT model for a specific language pair, the translations used for training and evaluation may contain inaccuracies, although these translations have been assessed and rectified if necessary by human annotators.

We do not implement the operations highlighted in red, introduced in Section~\ref{subsec:compass_ops}; their actual implementation and integration into ConvXAI systems are left for future work.

Given the difficulties involved in recruiting multiple annotators - especially for low-resource languages - and considering the relatively straightforward nature of the annotation task, we limit our reporting of inter-annotator agreement (IAA) to only the German and Chinese test sets (\S\ref{subsec:dataset_construction}).

We do not extensively experiment with every model from different model families; rather, we select three widely used models of varying sizes (\S\ref{sec:model}).

The current state-of-the-art ConvXAI systems are typically designed to support a set of representative and widely used XAI methods, from which we determined the current set of 11 XAI approaches. Furthermore, extending the set of XAI operations is a highly involved process -- for example, it requires collecting user questions, translating them from English into all target languages, conducting user studies to evaluate translation quality, and recruiting annotators (which is particularly challenging for low-resource languages, such as Telugu in our case).

\section*{Ethics Statement}
The participants in the machine translation evaluation were compensated at or above the minimum wage,
in accordance with the standards of our host institutions’
regions. The annotation took each annotator approximately 8 hours on average.

\section*{Acknowledgment} 
We thank Alon Drobickij and Selin Yeginer for reviewing the German translation, Polina Danilovskaia for reviewing the Russian translation and Ravi Kiran Chikkala for reviewing the Telugu translation. We would extend our gratitude to Lisa Raithel for setting up and organizing the translation quality check for the German translation. 

Additionally, we are indebted to the anonymous reviewers of EMNLP 2025 for their helpful and rigorous feedback.
This work has been supported by the Federal Ministry of Research, Technology and Space (BMFTR) as part of the projects BIFOLD 24B, TRAILS (01IW24005), VERANDA (16KIS2047) and newspolygraph (03RU2U151C).

\bibliography{custom}

\appendix


\section{The \data{CoXQL} Dataset}
\label{app:coxql}
\subsection{Operations}
\input{table/coxql_ops}
Table~\ref{tab:coxql_ops} demonstrations all operations supported in \data{CoXQL}.

\subsection{Examples}
\label{app:coxql_examples}
\begin{figure*}[t!]
\centering
\resizebox{\textwidth}{!}{
\begin{minipage}{\columnwidth}
\includegraphics[width=\columnwidth]{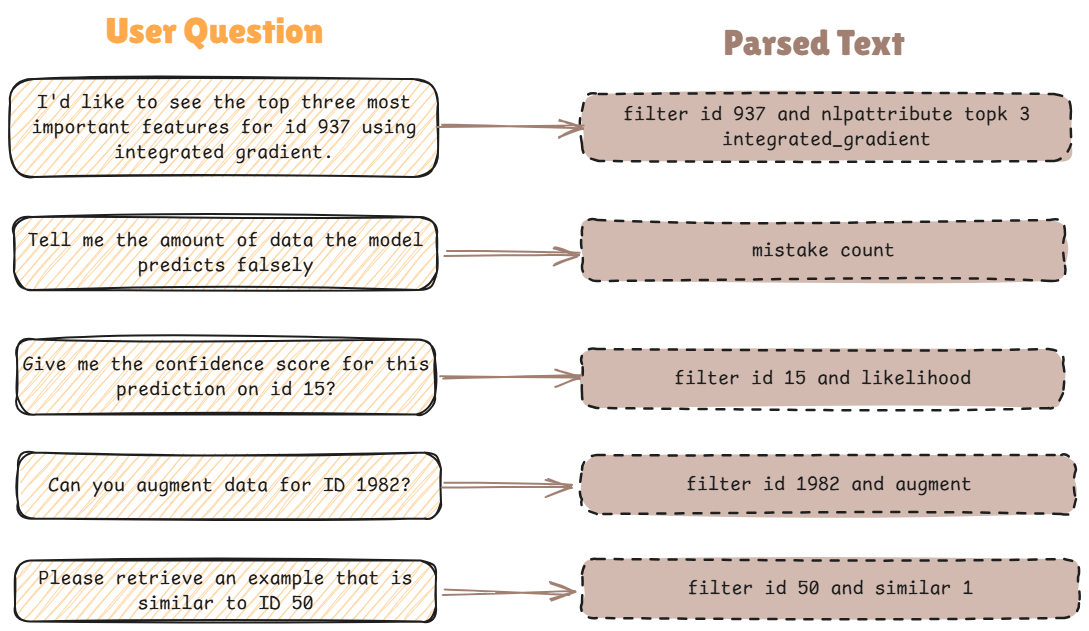}
\end{minipage}
}
\caption{Examples in \data{CoXQL} consist of user questions along with their corresponding parsed text, which are used to address various tasks such as feature importance, identifying mistakes, likelihood analysis, data augmentation, and instance similarity.}
\label{fig:coxql_example}
\end{figure*}

Figure~\ref{fig:coxql_example} displays the examples from the \data{CoXQL} dataset.

\section{Prompt Instruction for Machine Translation}
\label{app:prompt_instruction}
Figure~\ref{fig:translation} shows the prompt instruction used to perform the machine translation with \lm{Gemini-1.5-pro} for the \data{MultiCoXQL} dataset.
\begin{figure*}
    
    \begin{tcolorbox}[colback=orange!10!white, colframe=orange!55!black, title=Prompt Instruction]
    system\_prompt = "You are an excellent translator."

task\_instruction = f"Please translate the following text into \{language\}. Provide only the translated texts: \{original\_input\}

prompt = f"\{system\_prompt\} \{task\_instruction\}"
    \end{tcolorbox}

    \caption{Prompt instruction for machine translation.
    }
    \label{fig:translation}
\end{figure*}




\section{Guided Multi-prompt Parsing}
\label{app:guided_mp}

\begin{figure*}[h!]
    \centering
    \includegraphics[width=0.8\linewidth]{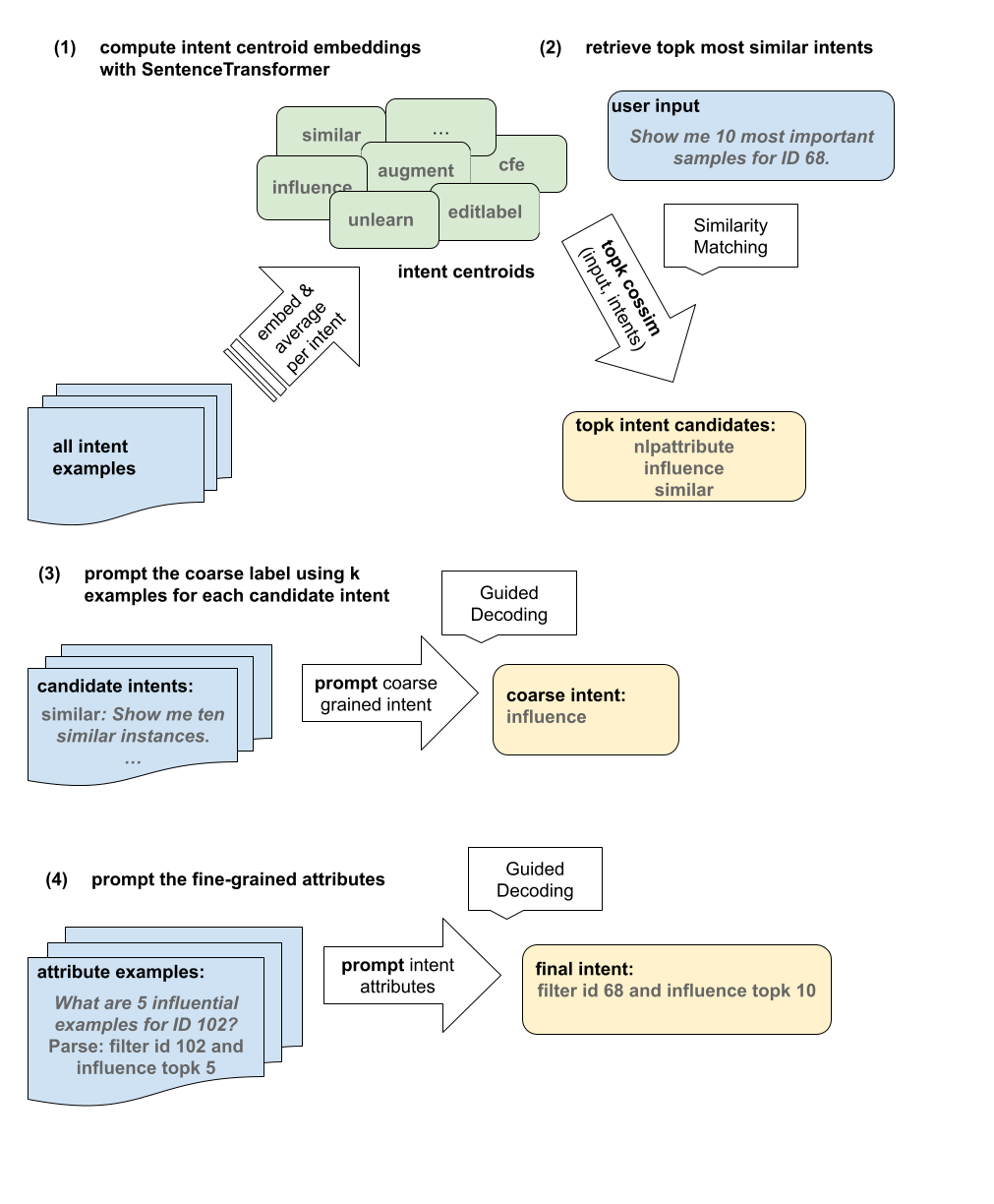}
    \caption{Pipeline of Guided Multi-prompt Parsing approach.}
    \label{fig:guided_mp_workflow}
\end{figure*}

Figure \ref{fig:guided_mp_workflow} illustrates the workflow of the Guided Multi-prompt Parsing (GMP) approach. This method combines the strengths of the Multi-prompt Parsing \cite{wang-etal-2024-llmcheckup} and Guided Decoding \cite{slack-2023-talktomodel} that proved to be effective for explanation request parsing in ConvXAI systems on the English data, but achieve substantially worse performance on other languages (Table \ref{tab:multi_coxql_parsing}).

First, GMP computes centroid embeddings using a multilingual SentenceTransformer model\footnote{\url{https://huggingface.co/sentence-transformers/paraphrase-multilingual-MiniLM-L12-v2}} for each intent (step (1) in Figure~\ref{fig:guided_mp_workflow}). Next, the user input is encoded with the same model and GMP retrieves the top $k$ most similar intents based on the cosine similarity between their centroid embeddings and the user query in step (2). In this way, we can have a selection of multiple candidate intents that are similar to the user query, but we are not restricted to a single most similar intent (e.g., \textit{nlpattribute, influence, etc.} can be chosen as candidates for the user question \textit{``Show me 10 most important samples for ID 68.''}). While similarity-based intent selection is also used in Guided Decoding \cite{slack-2023-talktomodel}, the key difference is that GMP uses the retrieved candidate intents to dynamically construct a prompt in step (3) that includes demonstrations for each of the candidates. Meanwhile, GMP excludes any dissimilar intents, so that the prompt is more concise and relevant to the input. GMP also uses a simplified intent-only grammar with guided decoding to make sure that the generated labels are from a pre-defined set. Note that the simplified grammar does not specify any attributes, only the main XAI operation (e.g. \textit{influence}).

Finally, in step (4), GMP refines the initial coarse-grained intent annotation and prompts the model with more examples for the selected intent to fill in the missing attributes. This step is similar to Multi-prompt Parsing \cite{wang-etal-2024-llmcheckup}, but instead of relying on a single grammar that covers all operations and their attributes, GMP uses an intent-specific grammar based on the selected intent from step (3) to ensure that we do not generate any attributes that are not valid for the selected XAI operation.

GMP is a flexible approach that leverages the advantages of multi-stage prompting that iteratively refines the predictions based on relevant demonstrations and guided decoding that constrains generated outputs. Thus, GMP generally achieves the best results on the \data{MultiCoXQL} dataset in the multilingual setting (Table~\ref{tab:multi_coxql_parsing}). 

\section{Sample Dataset Examples}
\label{app:sample_dataset}

Figure~\ref{fig:samples} presents examples of datasets (\S\ref{subsec:dataset_construction}) from which custom inputs are collected.

\begin{figure*}[h!]
    
    \centering

    \begin{tcolorbox}[colback=green!10!white, colframe=green!55!black, title=ECQA (Commonsense Question Answering)]
    \textbf{Question}: He had a lot on his plate opening business, this cause a lot of what?

    \textbf{Choices}: headaches, making money, success, failure, stress
    \end{tcolorbox}

    \begin{tcolorbox}[colback=green!10!white, colframe=green!55!black, title=COVIDFact (Medical Fact Checking)]

    \textbf{Claim}: Measuring sars-cov-2 neutralizing antibody activity using pseudotyped and chimeric viruses
    
    \textbf{Evidence}: While each surrogate virus exhibited subtle differences in the sensitivity with which neutralizing activity was detected, the neutralizing activity of both convalescent plasma and human monoclonal antibodies measured using each virus correlated quantitatively with neutralizing activity measured using an authentic SARS-CoV-2 neutralization assay. Here, we describe a collection of approaches based on SARS-CoV-2 spike-pseudotyped, single-cycle, replication-defective human immunodeficiency virus type-1 (HIV-1) and vesicular stomatitis virus (VSV), as well as a replication-competent VSV/SARS- CoV-2 chimeric virus.
    \end{tcolorbox}

    \begin{tcolorbox}[colback=green!10!white, colframe=green!55!black, title=SST2 (Sentiment Analysis)]
    \textbf{Review:} Allows us to hope that nolan is poised to embark a major career as a commercial yet inventive filmmaker.
    \end{tcolorbox}

    \caption{Examples of \data{ECQA}, \data{COVIDFact}, and \data{SST2} datasets, from which custom inputs are collected.
    }
    \label{fig:samples}
\end{figure*}

\section{Custom Input Extraction}
\label{app:custom_input_prompt}
Figure~\ref{fig:first_prompts} and Figure~\ref{fig:second_prompts} show the prompt instructions for \texttt{Na\"ive}, \texttt{TANL}, \texttt{GPT-NER} and \texttt{GOLLIE} approaches.

\begin{figure*}[h!]
    
    \centering

    \begin{tcolorbox}[colback=purple!50!white, colframe=purple!90!gray, title=Na\"ive]
    You will be given a user question related to explainability. Your task is to identify and extract the custom input from this question. The custom input refers to the specific information provided by the user that is necessary to fulfill their request. Extracting this input is crucial for processing user questions and taking appropriate actions. Please return only the custom input as a text string. If no custom input is clearly present, return an empty string. Below are some examples:
    
    \bigskip
    \textbf{[User Question]} {user question} 
    \bigskip
    
    \textbf{[Custom Input]} {custom input}
    \end{tcolorbox}

    \begin{tcolorbox}[colback=purple!50!white, colframe=purple!90!gray, title=TANL]
    You will be given a user question related to explainability. Your task is to identify and extract the custom input from this question. The custom input refers to the specific information provided by the user that is necessary to fulfill their request. Extracting this input is crucial for processing user questions and taking appropriate actions. Use the format `[ extracted\_text | custom\_input ]` to annotate the custom input in the output. Please return a text string with the custom input marked with [ extracted\_text | custom\_input ]. If no custom input is clearly present, return an empty string. Below are some examples:
    
    \bigskip
    
    \textbf{[User Question]} {user question} 
    \bigskip
    
    \textbf{[Custom Input]} {custom input}
    \end{tcolorbox}

    \caption{The prompt instructions for \texttt{Na\"ive} and \texttt{TANL} in custom input extraction.
    }
    \label{fig:first_prompts}
\end{figure*}

\begin{figure*}[h!]
    
    \centering

    \begin{tcolorbox}[colback=purple!50!white, colframe=purple!90!gray, title=GPT-NER]
    You are an excellent linguist. You will be given a user question related to explainability. The task is to label the custom input in the given user question. The custom input refers to the specific information provided by the user that is necessary to fulfill their request. Extracting this input is crucial for processing user questions and taking appropriate actions. Use special tokens @@\#\# to mark the extracted phrase in your response. Please return a text string with the custom input marked with @@\#\#. If no custom input is clearly present, return an empty string. Below are some examples:
    
    \bigskip
    
    \textbf{[User Question]} {user question} 
    \bigskip
    
    \textbf{[Custom Input]} {custom input}
    \end{tcolorbox}

    \begin{tcolorbox}[colback=purple!50!white, colframe=purple!90!gray, title=GOLLIE]
    You will be given a user question related to explainability. Your task is to identify and extract the custom input from this question. Please return a list of custom input.If no custom input is clearly present, return an empty list.Below is the schema for the custom input annotation:
    \bigskip
    
    @dataclass\\
    class CustomInput(Entity):\\
        """\\
        The custom input refers to the specific information provided by the user that is necessary to fulfill their request.\\
        Extracting this input is crucial for processing user questions and taking appropriate actions.\\
        """
    
    \bigskip
    
    \textbf{[User Question]} {user question} 
    \bigskip
    
    \textbf{[Custom Input]} {custom input}
    \end{tcolorbox}

    \caption{The prompt instructions for \texttt{GPT-NER} and \texttt{GOLLIE} in custom input extraction.
    }
    \label{fig:second_prompts}
\end{figure*}

\section{Models}
\label{app:model}
\subsection{Pre-trained \lm{BERT}-type Models}
Table~\ref{tab:bert_models} lists detailed information about used \lm{(m)BERT} models in our experiments. Fine-tuning \lm{(m)BERT} models for monolingual, cross-lingual, and multilingual evaluations can be completed within 20 minutes.

\input{table/bert_models}

\subsection{Decoder-only LLMs}
\input{table/models}
Table~\ref{tab:used_model} presents details of the three LLMs used in our experiments (\S\ref{sec:model}), including model sizes and corresponding URLs from the Hugging Face Hub. All models were directly obtained from the Hugging Face repository. All experiments were conducted using A100 or H100 GPUs. For each model, experiments on \data{MultiCoXQL} can be completed within 15 minutes.  For each model, experiments on \data{Compass} can be completed within 20 minutes.

\section{Human Evaluation Instructions for Translation Quality}
\label{app:human_evaluation}
\begin{figure*}[t!]
\centering
\resizebox{\textwidth}{!}{
\begin{minipage}{\columnwidth}
\includegraphics[width=\columnwidth]{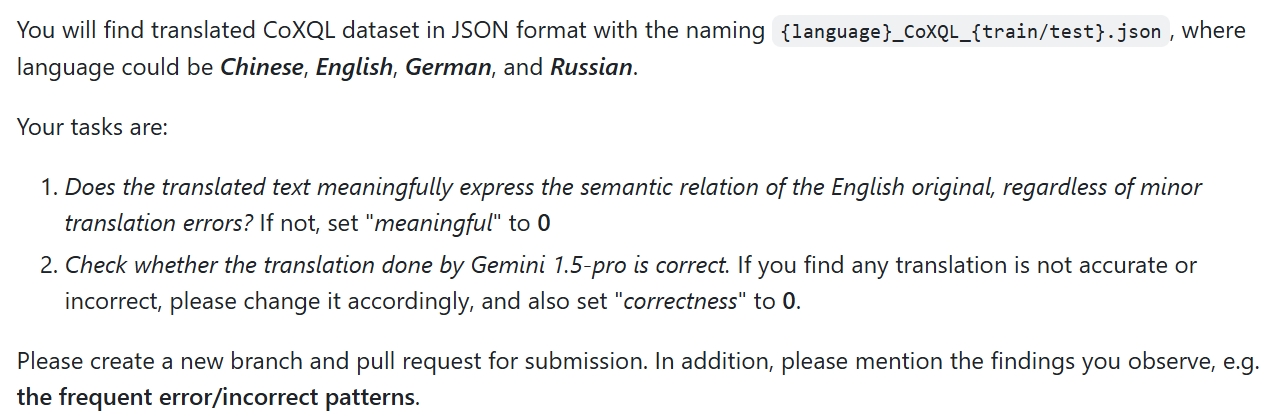}
\end{minipage}
}
\caption{Instructions for human evaluation given to human annotators.}
\label{fig:human_instruction}
\end{figure*}

Figure~\ref{fig:human_instruction} presents the instructions for human evaluation, which are used to guide annotators in assessing the quality of translations. All participants have a background in computational linguistics or computer science, hold at least a bachelor's degree, and are proficient in English. In addition, they are native speakers of one of the target languages (\S\ref{subsec:automatic_translation}).

\begin{figure*}[t!]
\centering
\resizebox{\textwidth}{!}{
\begin{minipage}{\columnwidth}
\includegraphics[width=\columnwidth]{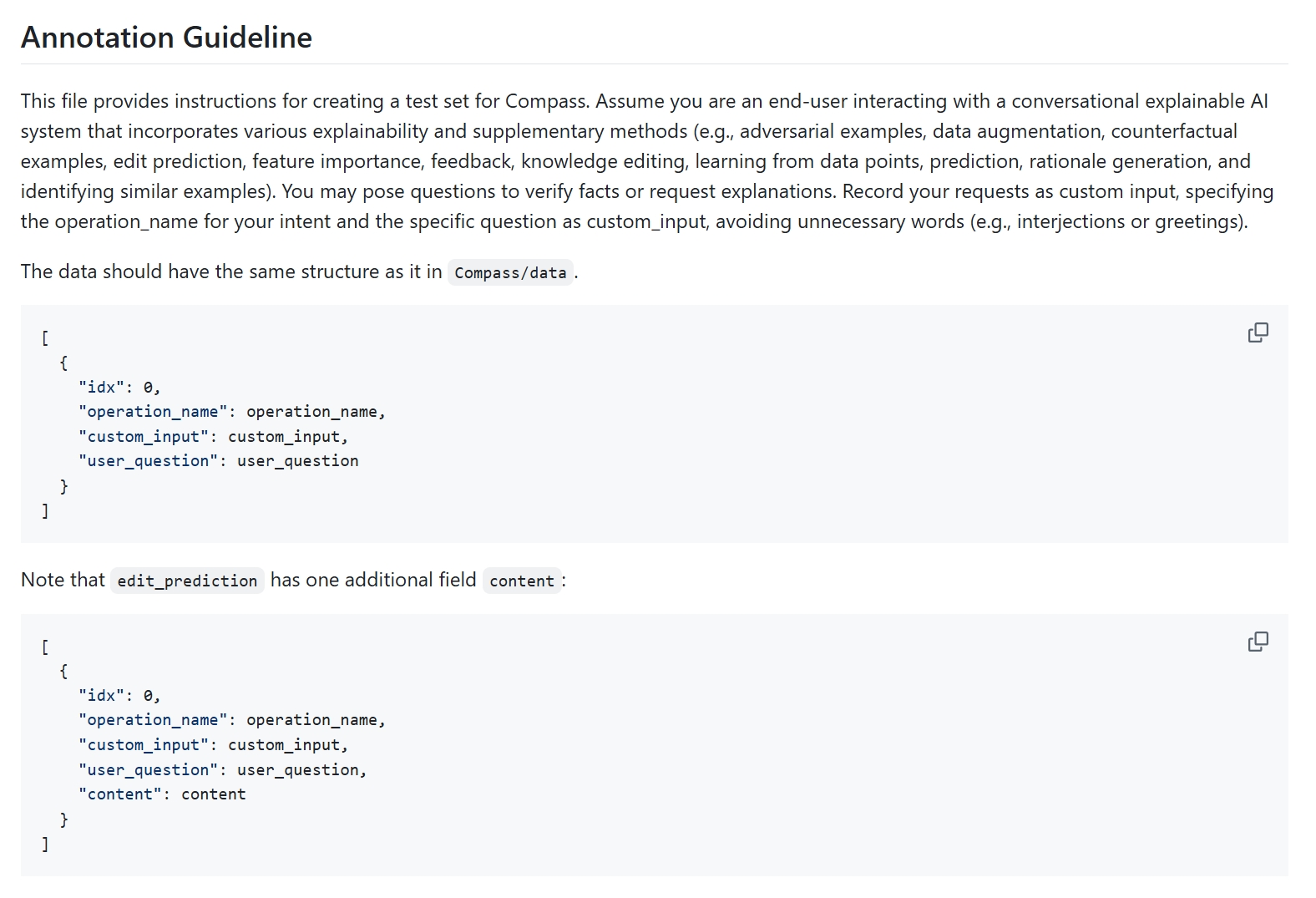}
\end{minipage}
}
\caption{The annotation guideline for human annotators in creating the \data{Compass} test set.}
\label{fig:testset}
\end{figure*}

\section{Annotation Guideline for Creating the \data{Compass} Test Set}
\label{app:testset}
Figure~\ref{fig:testset} shows the annotation instruction for creating the \data{Compass} test set.

\section{Translation Quality Analysis}
\label{app:quality_analysis}
\subsection{Chinese Translation}
For the \underline{\textit{Chinese}} translation, we found that \lm{Gemini-1.5-pro} sometimes omits prepositions. For example, \textit{``For id 9, what are the other 3 instances that are similar to it?''} is translated as \begin{CJK*}{UTF8}{gbsn}``ID 为 9 的条目，还有哪 3 个类似的实例?''\end{CJK*}, where the preposition \textit{``for''} is missing. Meanwhile, some words, which could have multiple meanings, such as \textit{``item''}, may be translated to a meaning that does not fit our context (\begin{CJK*}{UTF8}{gbsn}``商品''\end{CJK*} - \textit{``commodity/merchandise''}). Additionally, domain-specific terms, such as \textit{``accuracy score''} are often translated literally (\begin{CJK*}{UTF8}{gbsn}``得分准确率''\end{CJK*}) rather than using the correct predefined terminology (\begin{CJK*}{UTF8}{gbsn}``准确率评分''\end{CJK*}).


\subsection{Russian Translation}
For the \underline{\textit{Russian}} translation, we found three different categories of errors. The first category corresponds to the lack of context and ambiguous terms, e.g. \textit{``gold labels''} can be translated into \textit{``gold label stickers''} (``\rus{золотые этикетки}'') and \textit{``item ids''} into \textit{``product identifiers''} which is an acceptable translation but in a different setting. For instance, \textit{``Can you show me the item IDs in the training data?''} was translated into ``\rus{Вы можете показать мне идентификаторы товаров в обучающих данных?}'' This category of errors also includes ambiguous terms or terms with multiple possible meanings. E.g., both precision and accuracy can be translated as ``\rus{точность}'' in Russian but in our setting they should refer to two different metrics. 

The second category relates to the domain-specific terminology and abbreviations. For instance, NLP is sometimes directly transliterated as ``\rus{НЛП}'' which is used in Russian as an abbreviation for \textit{``Neural Linguistic Programming''}, not \textit{``Natural Language Processing''}. Also, \textit{``adversarial examples''} are frequently mistranslated as ``\rus{противоборствующие примеры}'' instead of a commonly used term ``\rus{состязательные примеры}''. 

The third category includes all errors caused by the model's failure to correctly interpret the task. Sometimes the output is text in English saying \textit{``Please provide the text...''}. Interestingly, this happens most often for the rationalization operation examples, the model may get confused by the new ``instruction'' contained in the input and it tries to accomplish the task instead of doing a simple translation. E.g., for \textit{``offer a plain-English interpretation for id 201''} it outputs \textit{``Please provide the text you would like me to translate. I need the text to be able to translate it to Russian and offer an interpretation for id 201.''}.

Additionally, the model frequently confuses instrumental and dative cases in Russian. It also misapplies adjective genders, using a single form when different genders are required. Pronoun coreference is often incorrect, leading to misinterpretations. Moreover, the model struggles with voice usage, incorrectly applying passive where active is needed and vice versa. It makes errors in verb aspect, confusing perfective and imperfective forms. Some translations sound unnatural due to weak word choices and direct, word-for-word rendering. Furthermore, the model sometimes applies English grammatical structures in a way that is ungrammatical in Russian.

\subsection{Telugu Translation}
In some instances, the system prompt translated into Telugu is included in the final translation, and there are cases where no direct equivalent exists for certain English words. For example, the word ``unlearn'' does not have an equivalent term in Telugu and is therefore replaced with a word meaning ``forget''. In other cases, the translation may be entirely different, with English words being substituted by phrases or terms that carry a similar meaning but are distinct in structure.

\subsection{German Translation}
Translation errors can occur in various forms, such as \textit{incorrect use of articles}, \textit{noun gender}, and \textit{case declination}, particularly when shorter substrings could be valid in different forms. Entire sentences may be omitted, especially if they resemble others in meaning within the same text, while idioms like "edge of your seat" or "keeping me guessing" are often translated literally instead of using equivalent expressions, leading to awkward or incorrect phrases in German (e.g.,``der Humor war flach'' instead of ``der Humor zündete nicht''). Other issues include splitting German compound verbs like ``angeben'' into ``geben ... an'' or omitting verbs in complex sentences, failing to adapt English words like ``all'' when a similar German word doesn’t fit the context, and even translating terms into antonyms, such as ``forgettable'' becoming ``unvergesslich''. Additionally, noun combinations may be mishandled (e.g., ``Sicherheits Ergebnisse'' instead of ``Sicherheitsergebnisse''), and while not outright wrong, translations can suffer from poor style—marked by excessive comma use that hampers readability and simpler, less elegant phrasing like ``von wo'' instead of ``woher''. These flaws often result in text that feels unnatural or unclear to native speakers, despite conveying the intended meaning.

\subsection{Semantic Similarity Comparison}
Table~\ref{tab:similarity} shows the semantic similarity between the original input
in English and the translated text in target languages. 

\input{table/similarity}

\section{\data{Compass} Dataset Translation}
The  prompt used for \lm{Gemini-1.5-pro} to translate texts from English to target languages is demonstrated in Figure~\ref{fig:translation_prompt}.

\begin{figure*}[t!]
    
    \centering

    \begin{tcolorbox}[colback=blue!20!white, colframe=gray!30!blue, title=Prompt for Machine Translation]
    The uploaded JSON consists of three fields: 
    user\_question, operation\_name, and 
    custom\_input. The custom\_input field is 
    derived from user\_question and serves as a 
    simplified version by discarding all 
    redundant information. Your task is to 
    translate both user\_question and 
    custom\_input into {language} while keeping 
    operation\_name as '{operation\_name}'. Note 
    that after the translation, the translated 
    custom\_input must remain a part of the 
    translated user\_question.

    \end{tcolorbox}
    
    \caption{The prompt used to translate user question and custom input into target languages for the \data{Compass} dataset.
    }
    \label{fig:translation_prompt}
\end{figure*}

\section{Error Analysis}

\subsection{Compass: Parsing}
\begin{figure*}[t!]
\centering
\resizebox{\textwidth}{!}{
\begin{minipage}{\columnwidth}
\includegraphics[width=\columnwidth]{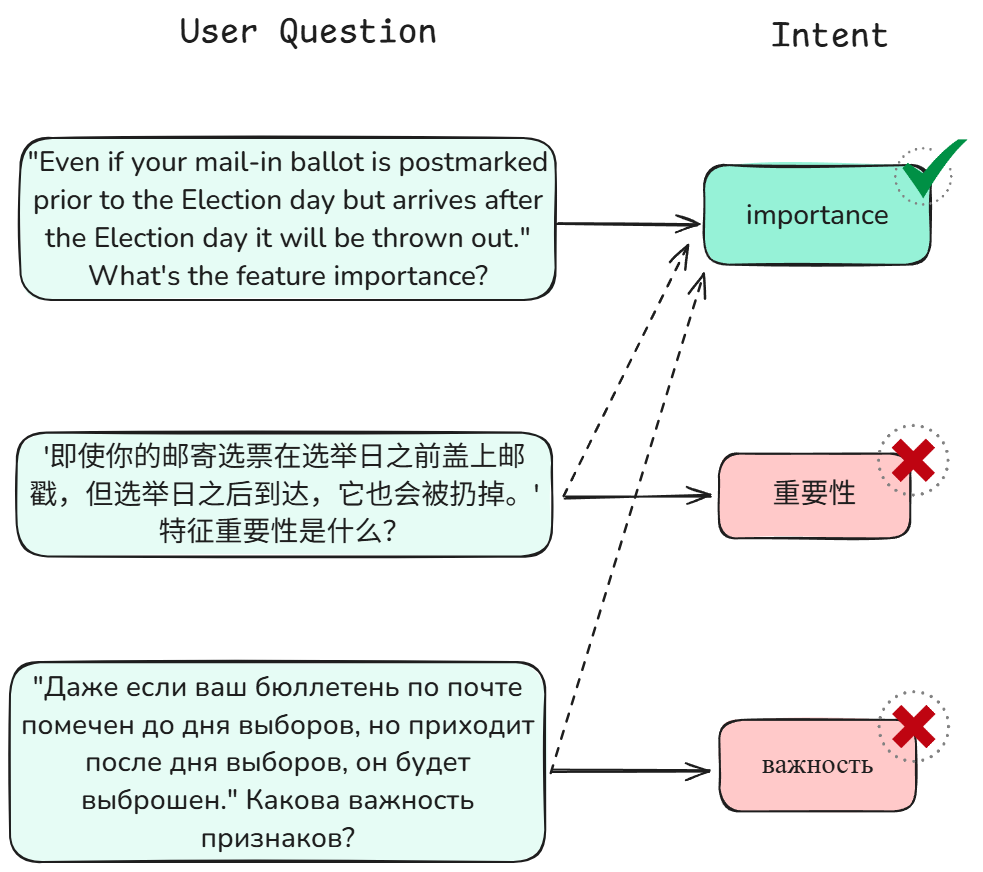}
\end{minipage}
}
\caption{The same example is shown in \textit{English}, \textit{Chinese}, and \textit{Russian}, along with their corresponding predicted intents. \textbf{Dashed} arrows indicate the \underline{ground-truth} label, while \textbf{solid} arrows represent the \underline{predicted} label.}
\label{fig:incorrect_example_intent}
\end{figure*}

Figure~\ref{fig:incorrect_example_intent} shows examples, where the LLMs generate labels (\texttt{importance}) in the target languages (\textit{Chinese} and \textit{Russian}) instead of in English.

\subsection{Compass: Custom Input Extraction}
\begin{figure*}[ht!]
\centering
\resizebox{\textwidth}{!}{
\begin{minipage}{\columnwidth}
\includegraphics[width=\columnwidth]{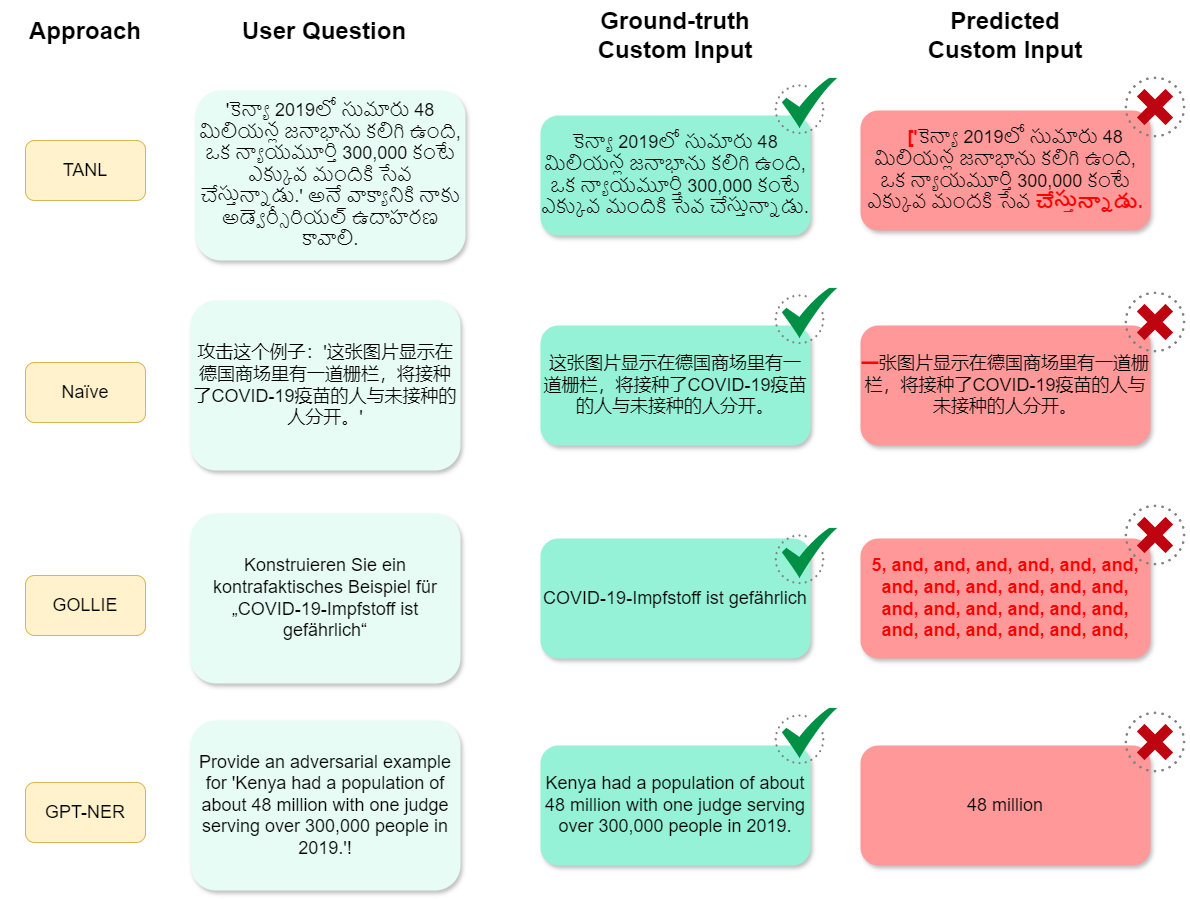}
\end{minipage}
}
\caption{Four example pairs in English, German, Chinese, and Telugu, each consisting of a user question, the corresponding ground-truth custom input, the predicted custom input, and the approach used. Redundant words or those not appearing in the user question are highlighted in \color{red}{red}.}
\label{fig:incorrect_example_custom_input}
\end{figure*}

Figure~\ref{fig:incorrect_example_custom_input} includes 4 example pairs in English, German, Chinese, and Telugu, each consisting of a user question, the corresponding ground-truth custom input, the predicted custom input, and the approach used (Na\"ive, TANL, GOLLIE, GPT-NER). Figure~\ref{fig:incorrect_example_custom_input} highlights several recurring mistakes in LLM-generated custom input extraction. Sometimes the models insert or replace terms that weren’t in the user’s original query, capture only a fragment of the desired input, or accidentally carry over artifacts from the extraction process into their output. There are also cases where, despite having ample examples to guide them, the LLMs simply fail to perform the extraction task. While Figure~\ref{fig:incorrect_example_custom_input} reveals that GOLLIE struggles with custom input extraction, it does not imply that GOLLIE is the worst-performing method; the error patterns described above are evident across nearly all of the evaluated approaches.

\end{document}

%% file: table/multicoxql_parsing.tex
\begin{table}[bt!]
    \centering
    \renewcommand*{\arraystretch}{1}

    \begin{minipage}{\columnwidth}
    \resizebox{\columnwidth}{!}{%
         \begin{tabular}{c|ccccc}

         \toprule

         \textbf{Approach} & \textbf{EN} & \textbf{ZH} & \textbf{DE} & \textbf{RU} & \textbf{TE}\\
         \midrule
         $\colorbox{purple}{NN}_{(Baseline)}$ & 44.25 & 44.25 & 40.71 & 42.48 & 25.66\\
        \bottomrule
         \end{tabular}
    }
    \end{minipage}

    \vspace{1em}
    \begin{minipage}{\columnwidth}
    \resizebox{\columnwidth}{!}{%

    \centering
        \begin{tabular}{cc|ccc|c}

        \toprule[1.5pt]
        \textbf{Model} & \textbf{Language}   & \colorbox{celeste}{GD} & \colorbox{brilliantlavender}{MP} & \colorbox{gblue}{MP+} & $\colorbox{orange}{GMP}_{(Ours)}$\\

        \midrule

        \multirow{5}{*}{\lm{Llama3}} & EN &    63.72 & \textbf{88.50} & 71.68 & 69.03\\
        & ZH &   58.41 & 43.36 & 51.33 & \textbf{72.57}\\
        & DE &   44.25 & 30.97 & 46.90  & \textbf{64.60}\\
        & RU &   48.67  & 43.46 & 52.21 & \textbf{71.68}\\
        & TE &   47.79 & 27.43 & 33.63 & \textbf{51.33} \\

        \midrule
        \multirow{5}{*}{\lm{Phi4}} & EN &   46.02  & 75.22  &  61.06  & \textbf{85.84}\\
        & ZH &   48.67 & 38.94 & 42.48 & \textbf{88.50}\\
        & DE &   38.94 & 30.97 & 30.97  & \textbf{78.76}\\
        & RU   &   40.71 & 26.55 & 39.20  & \textbf{84.96} \\
        & TE   &   53.98 & 14.16 & 14.16  & \textbf{77.88} \\

        \midrule
        \multirow{5}{*}{\lm{Qwen2.5}}& EN &   63.71 & 91.14 & \textbf{94.69}  & 88.50 \\
        & ZH &   68.14 & 55.75 & 57.52  & \textbf{88.50}\\
        & DE   &   59.29  & 46.02 & 48.67  & \textbf{77.88}\\
        & RU   &   69.03 & 58.41 & 64.60  & \textbf{85.84}\\
        & TE   &   63.72  & 40.71 & 42.48  & \textbf{77.88} \\

        \toprule[1.5pt]
        \end{tabular}

        }
    \end{minipage}
    \caption{
    Micro-$F_1$ scores (in \%) for different models on \data{MultiCoXQL} test set. \colorbox{purple}{NN} = Nearest Neighbor; \colorbox{celeste}{GD} = Guided Decoding prompted by 20-shots; \colorbox{brilliantlavender}{MP} = Multi-prompt Parsing; \colorbox{gblue}{MP+} = MP with template checks; \colorbox{orange}{GMP} = Guided Multi-prompt Parsing. \textbf{Bold-faced values} indicate the best-performing approach for a given language.
    }
    \label{tab:multi_coxql_parsing}
\end{table}

%% file: table/monolingual_intent_recognition.tex
\begin{table}[t!]
    \centering
    \footnotesize
    \renewcommand*{\arraystretch}{1}
    \resizebox{\columnwidth}{!}{%
        \begin{tabular}{c|ccccc}
            \toprule
            \textbf{Model} & \textbf{EN} & \textbf{ZH} & \textbf{DE} & \textbf{RU} & \textbf{TE}\\
            \midrule
             \lm{BERT} & 87.27 & \textbf{85.45} & \textbf{86.36} & 67.27 & 70.00 \\
             
             \lm{Llama3} & 84.55 & 50.91 & 65.45 & 60.00 & 53.64  \\
             \lm{Phi4} & 88.18 & 69.09 & 60.90 & 70.00 & 18.18  \\
             \lm{Qwen2.5} & \textbf{93.63} & 54.55 & 80.91 & \textbf{86.36} & \textbf{77.27} \\
            \bottomrule
        \end{tabular}
    }
    \caption{Micro-$F_1$ scores (in \%) on the \data{Compass} dataset are reported for the monolingual setting.}
    \label{tab:monoligual_intent_recognition}

\end{table}

%% file: table/multilingual_intent_recognition.tex
\begin{table}[tb!]
    \centering
    \footnotesize
    \renewcommand*{\arraystretch}{1}
    \resizebox{\columnwidth}{!}{%
        \begin{tabular}{c|ccccc|c}
            \toprule[1.5pt]
            \textbf{Data (\%)} & \textbf{EN} & \textbf{ZH} & \textbf{DE} & \textbf{RU} & \textbf{TE} & $\Delta$ \\
            \midrule

             - & 87.27 & 60.91 & 63.64 & 52.73 & 41.82 & - \\

             \midrule
             10\%  & 90.00 & 76.36 & 73.64 & 80.00 & 67.27 & 16.18\\
             25\% & 91.82 & 83.64 & 87.27 & 78.18 & 71.82 & 21.27 \\
             50\% & 90.00 & 82.73 & 83.64 & 85.45 & 77.27 & 22.54 \\ 
             75\% & 89.09 & 85.45 & 83.64 & 83.64 & 78.18 & 22.73 \\
             100\% & 91.82 & 86.36 & 92.73 & 86.36 & 84.55 & 27.09 \\
            \toprule[1.5pt]
        \end{tabular}
    }
    \caption{Micro-$F_1$ scores (in \%) on \data{Compass} for \textit{intent recognition} are reported in the multilingual setting, which are achieved by fine-tuning \lm{mBERT} on the complete English training split combined with different proportions of the translated target language training split, ranging from 10\% to 100\%. The final column shows the averaged improvement across languages compared to the cross-lingual evaluation.}
    \label{tab:multiligual_intent_recognition}

\end{table}

%% file: table/custom_input_extraction.tex
\begin{table}[t!]
    \centering
    \footnotesize
    \renewcommand*{\arraystretch}{1}
    \resizebox{\columnwidth}{!}{%
        \begin{tabular}{cc|cccc}
            \toprule[1.5pt]
            \multirow{2}{*}{\textbf{Language}} & \multirow{2}{*}{\textbf{Model}}& \multicolumn{4}{c}{{\textbf{Approaches}}}\\
            &  & \small{\texttt{Na\"ive}} & \small{\texttt{GPT-NER}} & \small{\texttt{TANL}} & \small{\texttt{GOLLIE}} \\

            \midrule
            
            \multirow{4}{*}{\raisebox{-0.1\height}{\includegraphics[width=0.3cm]{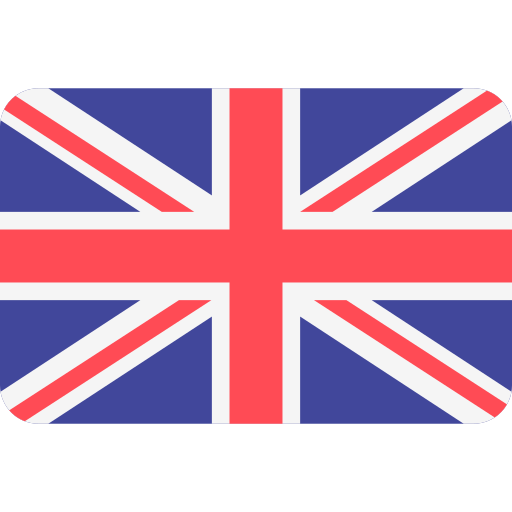}} \textbf{EN}} & \lm{BERT} & 71.96 & - & - & -  \\
             & \lm{Llama3-8B} & 64.55 & 58.18 & 60.91 & \textbf{66.36}\\
             & \lm{Phi4-14B} & \textbf{85.45} & 44.55 & 44.93 & 77.27\\
             & \lm{Qwen2.5-72B} & \textbf{89.09} & 77.27 & 68.18 & 80.91\\

             \midrule
            
            \multirow{4}{*}{ \raisebox{-0.1\height}{\includegraphics[width=0.3cm]{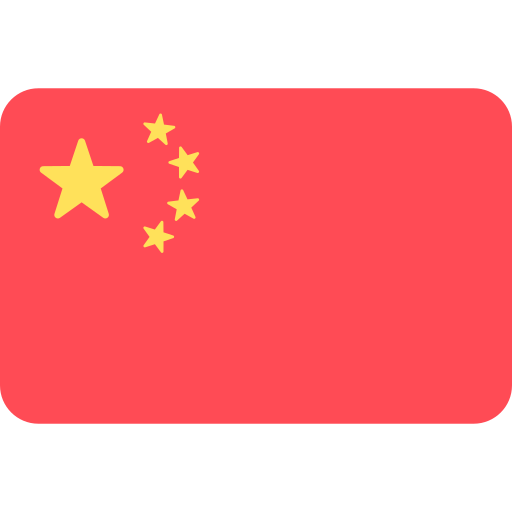}} \textbf{ZH}} & \lm{BERT} & 69.82 & - & - & - \\
             & \lm{Llama3-8B} & 20.91 & 60.91 & 48.18 & \textbf{62.73}\\
             & \lm{Phi4-14B} & 26.36 & 50.00 & 32.73 & \textbf{73.63} \\
             & \lm{Qwen2.5-72B} & 45.45 & \textbf{70.00} & 50.91 & 68.18 \\
            
             \midrule
            
            \multirow{4}{*}{\raisebox{-0.1\height}{\includegraphics[width=0.3cm]{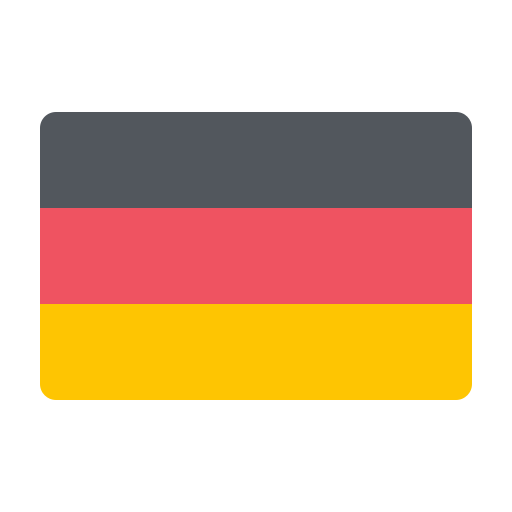}} \textbf{DE}} & \lm{BERT} & 76.69 & - & - & - \\
             & \lm{Llama3-8B} & 49.09 & 57.27 & 52.73 & \textbf{67.27}\\
             & \lm{Phi4-14B} & 52.73 & 40.00 & 38.18 & \textbf{72.73}\\
             & \lm{Qwen2.5-72B} & \textbf{77.27} & 60.91 & 48.18 & 70.91\\

             \midrule
            
           \multirow{4}{*}{\raisebox{0.1\height}{\includegraphics[width=0.3cm]{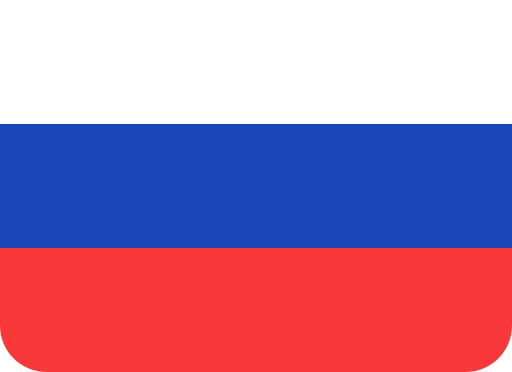}} \textbf{RU}} & \lm{BERT} & 74.95 & - & - & -\\
             & \lm{Llama3-8B} & 54.54 & 62.73 & 60.91 & \textbf{73.64}\\
             & \lm{Phi4-14B} & 63.64 & 44.55 & 29.09 & \textbf{75.45}\\
             & \lm{Qwen2.5-72B} & \textbf{86.36} & 68.18 & 60.91 & 82.73\\
            
             \midrule
            
           \multirow{4}{*}{\raisebox{-0.1\height}{\includegraphics[width=0.3cm]{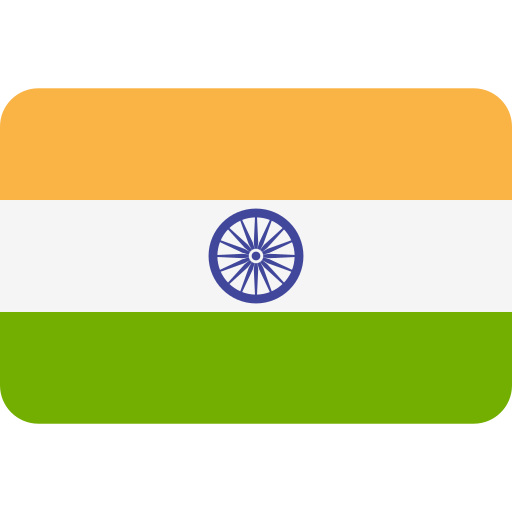}} \textbf{TE}} & \lm{BERT} & 78.38 & - & - & - \\
             & \lm{Llama3-8B} & \textbf{22.73} & 16.36 & 17.27 & 20.91\\
             & \lm{Phi4-14B} & 12.73 & 2.72 & 5.00 & \textbf{21.82}\\
             & \lm{Qwen2.5-72B} & \textbf{36.36} & 25.45 & 18.18 & 34.55 \\
            \toprule[1.5pt]  
        \end{tabular}
    }
    \caption{Custom input extraction results (Micro-$F_1$ scores in \%) obtained using \texttt{Na\"ive}, \texttt{GPT-NER}, \texttt{TANL}, \texttt{GOLLIE} on \data{Compass}, with \lm{BERT}, \lm{Llama3-8B}, \lm{Phi4-14B}, and \lm{Qwen2.5-72B}. Bold-faced values indicate the best-performing approach for a given LLM.}
    \label{tab:custom_input_extraction}

\end{table}

%% file: table/multilingual_custom_input_extraction.tex
\begin{table}[t!]
    \centering
    \footnotesize
    \renewcommand*{\arraystretch}{1}
    \resizebox{\columnwidth}{!}{%
        \begin{tabular}{c|ccccc|c}
            \toprule[1.5pt]
            \textbf{Data (\%)} & \textbf{EN} & \textbf{ZH} & \textbf{DE} & \textbf{RU} & \textbf{TE} & $\Delta$ \\
            \midrule

             - & 79.74 & 62.22 & 68.82 & 64.54 & 40.38 & - \\

             \midrule
             10\% & 80.13 & 80.32 & 75.49 & 78.32 & 80.72 & 15.86 \\
             25\% & 84.01 & 91.15 & 78.94 & 82.40 & 80.23 & 20.21 \\
             50\% & 83.19 & 92.00 & 79.23 & 83.56 & 82.06 & 20.87\\
             75\% & 84.15 & 92.87 & 80.18 & 84.16 & 83.42 & 21.82 \\
             100\% & 85.11 & 92.99 & 79.35 & 83.65 & 85.16 & 22.11 \\
            \toprule[1.5pt]
        \end{tabular}
    }
    \caption{Micro-$F_1$ scores (in \%) on \data{Compass} for \textit{custom input extraction} are reported in the multilingual setting, which are achieved by fine-tuning \lm{mBERT} on the complete English training split combined with different proportions of the translated target language training split, ranging from 10\% to 100\%. The final column shows the averaged improvement across languages compared to the cross-lingual evaluation.}
    \label{tab:multiligual_custom_input_extraction}

\end{table}

%% file: table/coxql_ops.tex
\begin{table*}[ht!]
    \centering
    \renewcommand*{\arraystretch}{1}
    \footnotesize
    \resizebox{\textwidth}{!}{%
        \roundedtable{p{0.2cm}p{5.3cm}|p{10cm}}{

        & \textbf{Operation} 
        & \textbf{Description/Request}
        \\
        
        \toprule
        \multirow{2}{*}{\centering \rotatebox[origin=c]{90}{\tiny{\textbf{Loc.Pr.}}}}
        & \texttt{\textbf{predict}(instance) } 
        & Get the prediction for the given instance 
        \\
        
        & \texttt{\textcolor{gray}{likelihood}(instance)} 
        & Calculate the model's confidence (or likelihood) on the given instance 
        \\

        \midrule 
        \multirow{2}{*}{\centering \rotatebox[origin=c]{90}{\tiny{\textbf{Glob.Pr.}}}}
        & \texttt{\textcolor{gray}{mistake}(\{sample|count\}, subset)} 
        & Count or show incorrectly predicted instances
        \\
        
        & \texttt{\textcolor{gray}{score}(subset, metric)} 
        & Determine the relation between prediction and labels
        \\

        \midrule 
        \multirow{2}{*}{\centering \rotatebox[origin=l]{90}{\tiny{\textbf{Loc. Expl.}}}} 
        & \texttt{\textbf{nlpattribute}(inst., topk, method) } 
        & Provide feature attribution scores 
        \\
        
        & \texttt{\textbf{rationalize}(inst.) } 
        & Explain the output/decision in natural language 
        \\
        
        & \texttt{\textcolor{gray}{influence}(inst., topk) } 
        & Provide the most influential training data instances 
        \\

        \midrule 
        \multirow{3}{*}{\centering \rotatebox[origin=c]{90}{\tiny{\textbf{Pertrb.}}}} 
        & \texttt{\textbf{cfe}(instance) } 
        & Generate a counterfactual of the given instance 
        \\

        & \texttt{\textbf{adversarial}(instance) } 
        & Generate an adversarial example based on the given instance 
        \\
        
        & \texttt{\textbf{augment}(instance) } 
        & Generate a new instance based on the given instance 
        \\

        \midrule 
        \multirow{4}{*}{\centering \rotatebox[origin=r]{90}{\tiny{\textbf{Data}}}}
        & \texttt{\textcolor{gray}{show}(instance)} 
        & Show the contents of an instance
        \\
        
        & \texttt{\textcolor{gray}{countdata}(list)} 
        & Count instances
        \\
        
        & \texttt{\textcolor{gray}{label}(dataset)} 
        & Describe the label distribution
        \\
        
        & \texttt{\textcolor{gray}{keywords}(topk)} 
        & Show most common words
        \\
        
        & \texttt{\textbf{similar}(instance, topk) } 
        & Show most similar instances
        \\

        \midrule 
        \multirow{3}{*}{\centering \rotatebox[origin=c]{90}{\tiny{\textbf{Mod.}}}} 
        & \texttt{\textcolor{gray}{editlabel}(instance)} 
        & Change the true/gold label of a given instance 
        \\
        
        & \texttt{\textbf{learn}(instance) } 
        & Retrain or fine-tune the model based on a given instance 
        \\
        
        & \texttt{\textcolor{gray}{unlearn}(instance) } 
        & Remove or unlearn a given instance from the model 
        \\

        \midrule
        \multirow{6}{*}{\centering \rotatebox[origin=c]{90}{\tiny{\textbf{Meta}}}}
        & \texttt{\textcolor{gray}{function}()} 
        & Explain the functionality of the system
        \\
        
        & \texttt{\textcolor{gray}{tutorial}(op\_name)} 
        & Provide an explanation of the given operation 
        \\
        
        & \texttt{\textcolor{gray}{data}()} 
        & Show the metadata of the dataset
        \\
        
        & \texttt{\textcolor{gray}{model}()} 
        & Show the metadata of the model 
        \\
        
        & \texttt{\textcolor{gray}{domain}(query)} 
        & Explain terminology or concepts outside of the system's functionality, but related to the domain
        \\
        
        \midrule
        
         \multirow{6}{*}{\centering \rotatebox[origin=r]{90}{\tiny{\textbf{Filter}}}}
        & \texttt{\textcolor{gray}{filter}(id)}
        & Access single instance by its ID 
        \\
        
        & \texttt{\textcolor{gray}{predictfilter}(label)}
        & Filter the dataset according to the model's predicted label
        \\
        
        & \texttt{\textcolor{gray}{labelfilter}(label)}
        & Filter the dataset according to the true/gold label given by the dataset
        \\
        
        & \texttt{\textcolor{gray}{lengthfilter}(level, len)}
        & Filter the dataset by length of the instance (characters, tokens, …)
        \\
        
        & \texttt{\textcolor{gray}{previousfilter}()}
        & Filter the dataset according to outcome of previous operation
        \\
        
        & \texttt{\textcolor{gray}{includes}(token)}
        & Filter the dataset by token occurrence
        \\

        \midrule 
        \multirow{2}{*}{\centering \rotatebox[origin=c]{90}{\textls[-50]{\tiny{\textbf{Logic}}}}}
        & \texttt{\textcolor{gray}{and}(op1, op2)} 
        & Concatenate multiple operations 
        \\
        
        & \texttt{\textcolor{gray}{or}(op1, op2)} 
        & Select multiple filters 
        \\
        }
        }
    \caption{
    Main operations in \data{CoXQL}, including exlainability (\textit{Local Explanation}, \textit{Perturbation}, \textit{Modification}) and supplementary (\textit{Local Prediction}, \textit{Global Prediction}, \textit{Data}, \textit{Meta}, \textit{Filter}, \textit{Logic}) operations. Operations designated in \textbf{bold} should facilitate custom input and, therefore, be selected for integration with \data{Compass} dataset.
    }
    \label{tab:coxql_ops}
\end{table*}

%% file: table/bert_models.tex
\begin{table*}[t!]
    \centering
    \renewcommand*{\arraystretch}{0.75}
    \resizebox{\textwidth}{!}{%
        \begin{tabular}{ccccc}

        \toprule
        \textbf{Name}& \textbf{Language} & \textbf{Citation} & \textbf{Size} & \textbf{Link}\\

        \midrule
        \lm{BERT} & English & \cite{devlin-etal-2019-bert} & 110M & \url{https://huggingface.co/google-bert/bert-base-uncased}\\

        \lm{BERT} & German & \cite{devlin-etal-2019-bert}& 110M & \url{https://huggingface.co/google-bert/bert-base-german-cased}\\

        \lm{BERT} & Chinese & \cite{devlin-etal-2019-bert}& 110M & \url{https://huggingface.co/google-bert/bert-base-chinese}\\

        \lm{BERT} & Russian & \cite{kuratov2019adaptationdeepbidirectionalmultilingual} & 110M & \url{https://huggingface.co/DeepPavlov/rubert-base-cased}\\

        \lm{BERT}& Telugu & \cite{joshi2023l3cubehindbertdevbertpretrainedbert} & 110M & \url{https://huggingface.co/l3cube-pune/telugu-bert}\\

        \lm{mBERT} & Multilingual & \cite{devlin-etal-2019-bert}& 110M & 
        \url{https://huggingface.co/google-bert/bert-base-multilingual-cased}\\
        
        \bottomrule
        \end{tabular}
        }
    \caption{
    Detailed information about used \lm{BERT} and \lm{mBERT} models in our experiments. 
    }
    \label{tab:bert_models}
\end{table*}

%% file: table/models.tex
\begin{table*}[t!]
    \centering
    \resizebox{\textwidth}{!}{%
        \begin{tabular}{cccc}

        \toprule
        \textbf{Name}& \textbf{Citation} & \textbf{Size} & \textbf{Link}\\

        \midrule
        
        \lm{Llama3} & \citet{llama3modelcard} & 8B &   \url{https://huggingface.co/meta-llama/Meta-Llama-3-8B}\\

        \lm{Phi4} &  \citet{abdin2024phi4technicalreport} & 14B & \url{https://huggingface.co/microsoft/phi-4}\\
        
        \lm{Qwen2.5} & \citet{qwen2024qwen25technicalreport} & 72B &   \url{https://huggingface.co/Qwen/Qwen2.5-72B}\\

        \bottomrule
        \end{tabular}
        }
    \caption{
    Detailed information about used LLMs in our experiments. 
    }
    \label{tab:used_model}
\end{table*}

%% file: table/similarity.tex
\begin{table}[t!]
    \centering
    \setlength{\extrarowheight}{3pt}
    \renewcommand*{\arraystretch}{0.8}
    
    \footnotesize
    \resizebox{\columnwidth}{!}{%
    \begin{tabular}{c|ccccc}
        \toprule
        \textbf{Dataset} & \textbf{Set} & \textbf{ZH} & \textbf{DE} & \textbf{TE} & \textbf{RU} \\
         \cline{1-6}
        \multirow{6}{*}{\centering \small{\rotatebox[origin=c]{90}{\textbf{\data{MultiCoXQL}}}}} & \multicolumn{5}{c}{\textit{\textbf{Before Correction}}} \\
        \cline{2-6}
        & \textit{Train} & 83.66\% & 84.18\% & 55.28\% & 82.74\% \\

        & \textit{Test} & 83.46\% & 85.36\% & 54.92\% & 83.21\% \\

        \cline{2-6}
        & \multicolumn{5}{c}{\textit{\textbf{After Correction}}} \\
         \cline{2-6}

        & \textit{Train} & 84.25\% & 85.83\%  & 53.12\% & 83.87\%\\

        & \textit{Test} & 82.66\% & 85.29\% & 54.58\% & 83.33\%  \\
        \hline

       \multirow{6}{*}{\centering \small{\rotatebox[origin=c]{90}{\textbf{\data{Compass}}}}} & \multicolumn{5}{c}{\textit{\textbf{Before Correction}}} \\
        \cline{2-6}
        & \textit{Train} & 85.75\% & 81.07\% & 37.56\% & 82.73\% \\

        & \textit{Test} & 84.51\% & 87.50\% & 70.14\% & 86.74\% \\

        \cline{2-6}
        & \multicolumn{5}{c}{\textit{\textbf{After Correction}}} \\
         \cline{2-6}

        & \textit{Train} & 85.97\% & 88.37\% & 37.86\% & 85.07\%\\

        & \textit{Test} & 84.51\% & 89.12\% & 73.59\% & 88.02\%  \\
        \bottomrule
    \end{tabular}
    }
    \caption{Semantic similarity between the original input in English and the translated texts in Chinese (ZH), German (DE), Telugu (TE) and Russian (RU) from the training and test sets of \data{MultiCoXQL} and \data{Compass} measured by a multilingual sentence transformer.}
    \label{tab:similarity}
\end{table}